\documentclass[conference]{IEEEtran}
\IEEEoverridecommandlockouts

\usepackage{cite}

\usepackage[hyphens]{url}
\usepackage{breakurl}
\usepackage{hyperref}
\hypersetup{breaklinks=true}

\usepackage{booktabs}

\usepackage{multirow}

\usepackage{amsmath,amssymb,amsfonts}
\usepackage{algorithmic}
\usepackage{graphicx}
\usepackage{textcomp}
\usepackage{fancyhdr}

\usepackage{xcolor}
\def\BibTeX{{\rm B\kern-.05em{\sc i\kern-.025em b}\kern-.08em
    T\kern-.1667em\lower.7ex\hbox{E}\kern-.125emX}}

\usepackage[caption=false,font=footnotesize]{subfig}

\usepackage{cleveref}

\begin{document}

\title{
Estimating the Diameter at Breast Height of Trees in a Forest from RGB
}

\author{%
  \IEEEauthorblockN{%
    Siming He\IEEEauthorrefmark{1}\IEEEauthorrefmark{2},\;
    Zachary Osman\IEEEauthorrefmark{1}\IEEEauthorrefmark{2},\;
    Fernando Cladera\IEEEauthorrefmark{2},\;
    Dexter Ong\IEEEauthorrefmark{2},\;
    Nitant Rai\IEEEauthorrefmark{3},\;\\
    Patrick Corey Green\IEEEauthorrefmark{3},\;
    Vijay Kumar\IEEEauthorrefmark{2},\;
    Pratik Chaudhari\IEEEauthorrefmark{2}%
  }
  \vspace{1ex} 
  \IEEEauthorblockA{%
    \IEEEauthorrefmark{1}Equal Contribution\\
    \IEEEauthorrefmark{2} General Robotics, Automation, Sensing and Perception (GRASP) Laboratory,
      University of Pennsylvania\\ 
    \IEEEauthorrefmark{3}Department of Forest Resources and Environmental Conservation,
      Virginia Tech\\[0.5ex]
      Email: \{siminghe, osmanz, fclad, odexter, kumar, pratikac\}@seas.upenn.edu,
    \{nitant3, pcgreen7\}@vt.edu
  }
}

\maketitle
\fancypagestyle{withfooter}{
  \renewcommand{\headrulewidth}{0pt}
  \fancyfoot[C]{\footnotesize Accepted to the Novel Approaches for Precision Agriculture and Forestry with Autonomous Robots IEEE ICRA Workshop - 2025}
}
\thispagestyle{withfooter}
\pagestyle{withfooter}

\begin{abstract}
Forest inventories rely on accurate measurements of the diameter at breast height (DBH) for ecological monitoring, resource management, and carbon accounting. While LiDAR‐based techniques can achieve centimeter‐level precision, they are cost‐prohibitive and operationally complex. We present a low‐cost alternative that only needs a consumer‐grade 360° video camera. Our semi‐automated pipeline comprises of (i) a dense point cloud reconstruction using Structure from Motion (SfM) photogrammetry software called Agisoft Metashape, (ii) semantic trunk segmentation by projecting Grounded Segment Anything (SAM) masks onto the 3D cloud, and (iii) a robust RANSAC‐based technique to estimate cross section shape and DBH. We introduce an interactive visualization tool for inspecting segmented trees and their estimated DBH. On 61 acquisitions of 43 trees under a variety of conditions, our method attains median absolute relative errors of 5–9\% with respect to ``ground-truth'' manual measurements. This is only 2–4\% higher than LiDAR-based estimates, while employing a single 360° camera that costs orders of magnitude less, requires minimal setup, and is widely available.
\end{abstract}


\section{Introduction}
\label{sec:introduction}

Forest ecosystems play a critical role in global carbon cycles
and biodiversity. Effective management and monitoring of
these resources require accurate and timely forest inventory
data. 
Diameter at Breast Height (DBH), typically measured at 1.3 meters above ground level, is one of the most
fundamental and frequently measured tree attributes in Forest
Resource Inventories (FRIs) \cite{f14091723, sahin, su16062275}. It serves as a key input for
estimating tree volume, biomass, carbon stock, and assessing
overall forest health and structure \cite{piponiot}. 

\begin{figure}[!t]
\centering
\includegraphics[width=\linewidth]{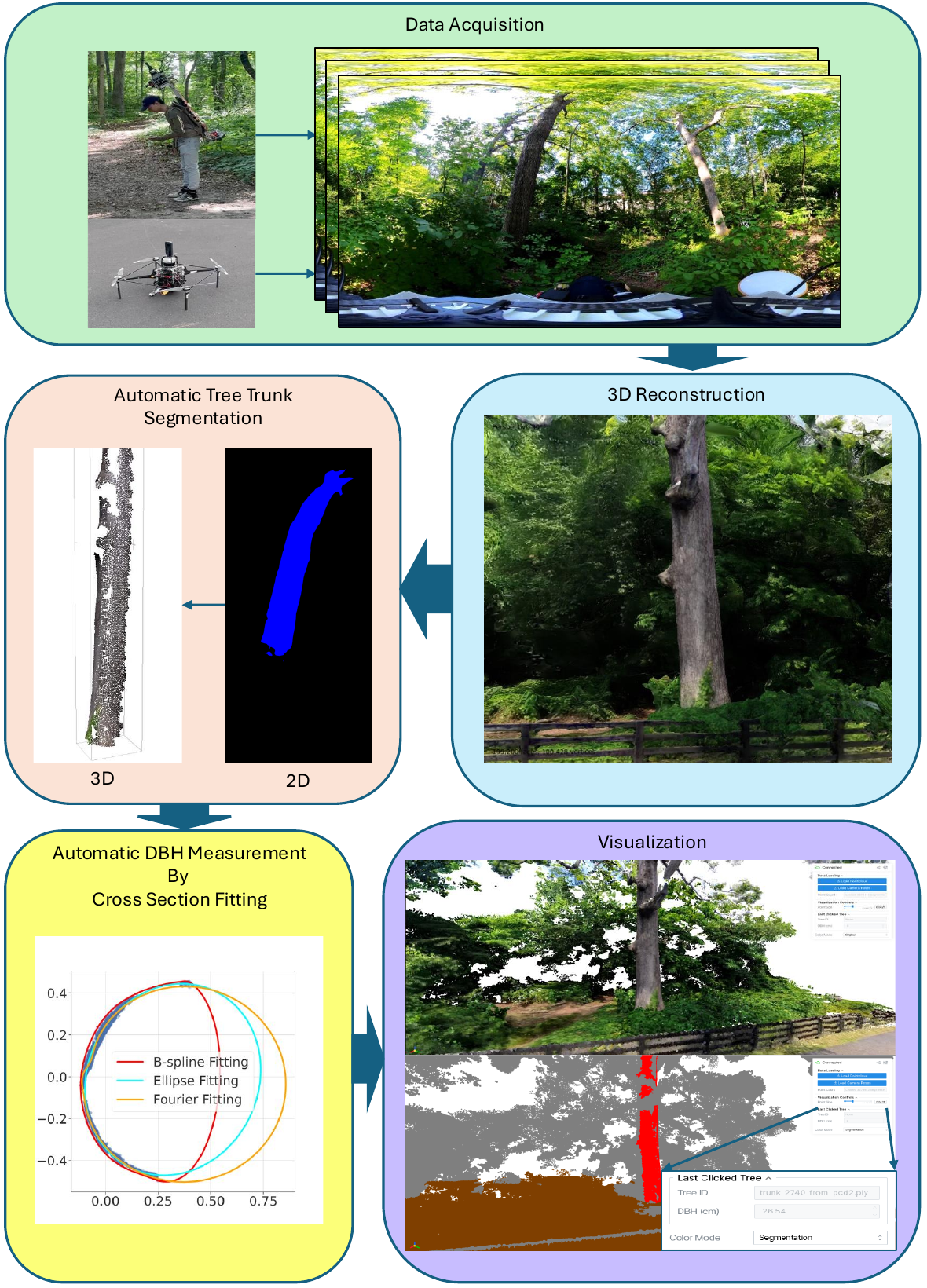} 
\caption{Overview of the proposed 360° camera-based DBH estimation pipeline. 360° video data is collected using a Falcon UAV and a backpack-mounted sensor tower. Agisoft Metashape is used to extract and align images, generate point cloud and mesh, and scale the model. By projecting Grounded SAM masks onto the point cloud, we can extract individual tree trunks. We estimate DBH using Fourier-based fitting with a RANSAC framework on cross sections at breast height. Using an interactive interface, we can visualize and inspect reconstructions and tree trunks with corresponding DBH estimates. }
\label{fig:pipeline}
\end{figure}

The advancement of precision agriculture and forestry,
driven partly by the challenges of climate change and the
need for sustainable practices, increasingly relies on automated
and efficient data acquisition methods \cite{babar2025sustainableprecisionagricultureinternet, sharma}. Autonomous robots
equipped with advanced sensing capabilities hold significant
potential for revolutionizing field data collection in these
domains. Accurate tree parameter estimation, including DBH, is explicitly highlighted as a key capability for these
robotic systems, enabling applications from carbon sequestration quantification to biomass estimation \cite{mattamala2024autonomousforestinventorylegged, karjalainen2025autonomousphotogrammetricforestinventory}.

Traditional DBH measurement methods, such as calipers and diameter tape, are easy to use but become highly inefficient and cumbersome for collecting data at scale \cite{weaver, f14050891, Song2021Handheld}. LiDAR offers state-of-the-art accuracy for DBH measurements, enabling simultaneous data collection from multiple trees with root mean squared errors (RMSE) often below 2 cm \cite{Gao2021TreeStructural, Zhou2019Diameter}. However, its high cost, reliance on specialized processing software, and need for trained personnel limit widespread adoption, driving interest in more accessible options \cite{morgenroth, Stoddart2023CCF}.

Photogrammetric methods based on Structure from Motion (SfM) present a compelling and cost-effective alternative to LiDAR.
SfM reconstructs 3D point clouds by analyzing sequences of overlapping 2D images, which can be captured using consumer-grade cameras, including omnidirectional (360°) systems \cite{su14095399,rs12060988,s24113534}. This technique has shown strong performance in forestry applications such as DBH estimation, achieving RMSE of 1–3 cm, comparable to LiDAR but at a significantly lower cost \cite{Piermattei, GaoQiang}. Nonetheless, SfM-derived point clouds are often noisier, less complete, and more susceptible to geometric distortions, particularly in densely vegetated or unevenly lit forest environments \cite{Iglhaut2019SfM, jaud}.

In this paper, we present a scalable, low-cost approach for estimating DBH using 360° video and SfM reconstruction. Our method leverages Agisoft Metashape, a commercially available photogrammetry software, along with a novel processing pipeline that incorporates automated trunk segmentation and robust DBH estimation. The key contributions of this work are as follows:
\begin{enumerate}
    \item A complete, semi-automated pipeline for transforming 360° video into DBH estimates and visualizations. The pipeline integrates SfM, deep learning-based semantic segmentation using Grounded SAM projection \cite{ren2024groundedsamassemblingopenworld}, and robust geometric fitting to enable accurate and efficient DBH estimation.
    \item A custom visualization tool for interactive inspection of segmented tree point clouds and associated DBH estimates, facilitating user validation and analysis.
    \item Extensive experiments on 61 samples of 43 trees shown in \cref{tab:data-summary,tab:performance-metrics}. Our experiments demonstrate that DBH estimation based on photogrammetry has higher accuracy, but lower precision than estimations based on LiDAR-inertial odometry. The photogrammetry method is 2-4\% worse than the per-LiDAR scan based method \cite{PRABHU2024111050}, but 360° cameras are significantly more affordable, easy to set up, and readily accessible. The comparison is reasonable since our paper and \cite{PRABHU2024111050} uses the same LiDAR and mobile platform and evaluates on trees of similar DBH range in similar forests.  Overall, our system is a viable alternative to LiDAR for large-scale or resource-constrained forestry applications. 
\end{enumerate}

The remainder of this paper is structured as follows: Section \ref{sec:related_work} reviews existing methods for DBH estimation and point cloud processing. Section \ref{sec:methodology} details our proposed pipeline. Section \ref{sec:experiments} presents the experimental setup and results, and discusses the findings, advantages, and limitations of the approach. Finally, Section \ref{sec:conclusion} outlines directions for future work.

\begin{figure}[!t]
\centering
\includegraphics[width=0.9\linewidth]{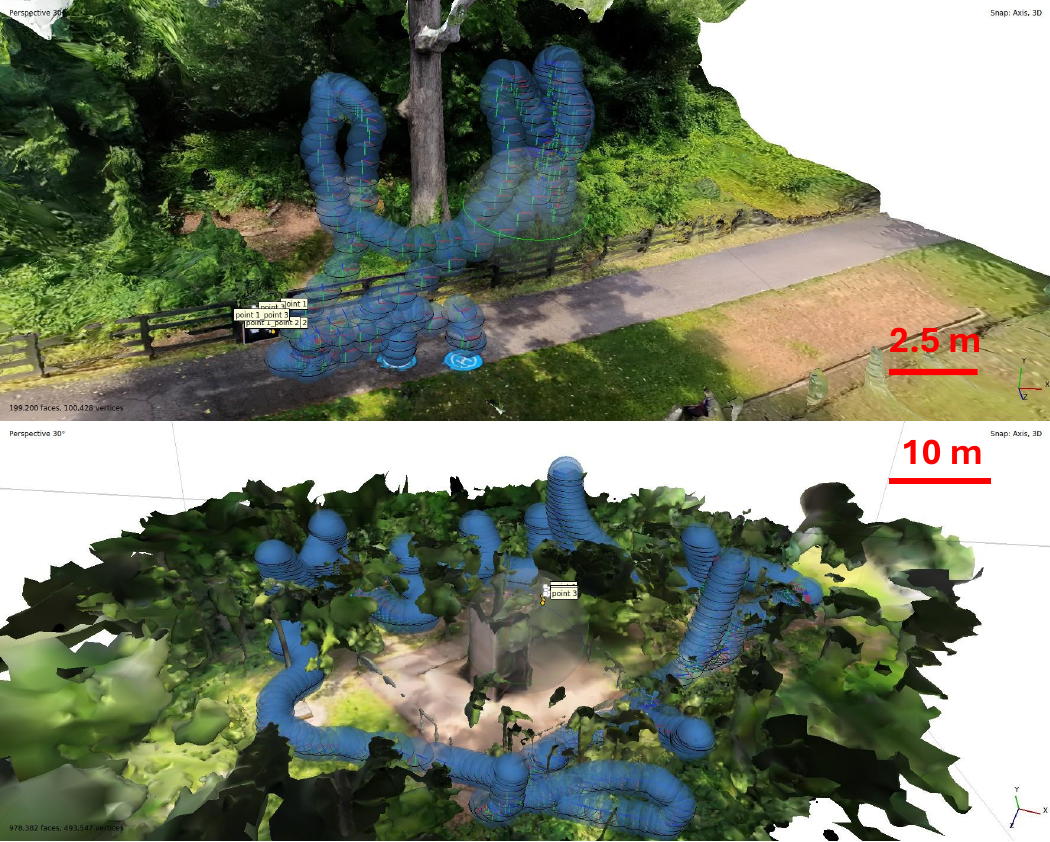} 
\caption{Examples of short and long trajectory sequences used for data collection. Blue spheres indicate camera positions over time. Scale bars are shown for reference. The short sequence captures a single tree at close range, while the long sequence covers a larger 60m$\times$60m area for multi-tree reconstruction.}
\label{fig:longshortreconstruction}
\end{figure}

\begin{table}[!t]
\label{tab:expsummary}
\centering
\caption{Summary of scanning experiments (61 samples of 43 trees).}
\label{tab:data-summary}
\begin{tabular}{l  c  c c}
\toprule
Scan Method & \# Samples & DBH (cm) & \# Species \\
\midrule
\begin{tabular}[c]{@{}l@{}}Flying drones\\with sensors\end{tabular}         & 20 & 15.24 – 121.92 & 7 \\
\begin{tabular}[c]{@{}l@{}}Walking with sensors\\on backpack\end{tabular}          & 41 & 15.24 – 121.92 & 7 \\
\bottomrule
\end{tabular}
\end{table}

\section{Related Work}
\label{sec:related_work}

\subsection{DBH Measurement Techniques}
Manual approaches for measuring DBH have long been the standard in forestry, providing reliable data for tree inventory and management. The most widely used tool is the diameter tape, a calibrated tape that converts circumference to diameter when wrapped around the trunk at 1.3 m above ground level, ensuring quick and accurate readings \cite{f14091723, wilsontools}. Other manual tools include tree calipers and the Biltmore stick, which provide direct diameter readings through visual alignment techniques \cite{wilsontools}. Standard protocols address various field complications, such as measuring on slopes, leaning trunks, trees with branches or bumps at breast height, and multi-stemmed or forked trees, by adjusting the measurement position or combining diameters according to established guidelines \cite{americanforests2019measuring}. These methods provide reliable data when performed carefully, but they are labor-intensive and time-consuming, requiring physical access to each tree and skilled operators to ensure accuracy, especially in plots with irregular or large trees.

LiDAR technology has significantly advanced DBH measurement in forestry, supporting both direct and indirect estimation methods across a range of platforms. Indirect approaches commonly use regression models that relate LiDAR-derived metrics—such as tree height, crown area, and crown ratio—to field-measured DBH, achieving strong predictive accuracy across species \cite{cao2013predicting, talmage2024diameter, ZHANG2023100089}. Area-based models further aggregate these metrics at the plot or stand level for broader assessments \cite{talmage2024diameter, ZHANG2023100089}. For direct estimation, Terrestrial Laser Scanning (TLS) systems produce high-resolution 3D point clouds, enabling automated stem segmentation and geometric fitting with extreme precision \cite{LIANG201663, CALDERS2020112102}. Airborne LiDAR, deployed from crewed aircraft or UAV, supports rapid, large-scale forest assessment using both individual tree and area-based approaches \cite{cao2013predicting, talmage2024diameter, ZHANG2023100089}. Recent advances in compact LiDAR hardware have enabled more accessible solutions, including smartphone-based systems and dedicated mobile applications, which offer practical, in-field DBH measurement, particularly for trees with regular stem forms \cite{Guenther31122024, howie, gulci2023measuring}. While these technologies offer promising solutions for rapid, low-cost forest measurements, their shorter range and variability compared to professional-grade systems can limit their accuracy and applicability in complex environments \cite{Magnuson}. In general, despite having high accuracy and efficiency in data collection, LiDAR systems remain costly, require specialized processing expertise, and are sensitive to canopy density and environmental conditions, which can lead to incomplete or noisy data and limit their suitability for routine fieldwork \cite{ciobotari2024lidardataacquisitionprocessing}.

SfM-based approaches have gained prominence as a low-cost and accessible alternative for DBH measurement in forestry. These methods typically rely on terrestrial image acquisition, using standard or fisheye lenses to capture overlapping photos around tree stems \cite{zeybek, mokros2020non}. The images are processed using SfM and Multi-View Stereo (MVS) to generate detailed 3D point clouds, from which DBH is estimated by slicing the point cloud at breast height and fitting geometric shapes, such as circles or ellipses, to the stem cross-section. Both automated and semi-automated pipelines rely on careful post-processing of the SfM point cloud before extracting DBH estimates \cite{zeybek, GaoQiang}. SfM has also been applied to aerial imagery captured by drones, which enables the reconstruction of forest structures at larger scales. Although direct DBH measurement from aerial views is difficult, DBH can be estimated indirectly using relationships derived from tree height or crown dimensions \cite{SWAYZE2022101729, iizuka}. Other methods utilize stereoscopic imaging to acquire data, allowing for efficient reconstruction and automated extraction of breast height and stem geometry \cite{Eliopoulos_2020}. A notable advantage of many SfM-based techniques is their use of consumer-grade cameras and smartphones, making them an appealing option for large-scale or resource-constrained applications. However, the accuracy
and quality of SfM reconstructions are highly dependent
on image quality, which can be influenced by uncontrollable
environmental factors such as lighting conditions, surface
texture, and vegetation density \cite{Iglhaut2019SfM, nielsen}.

\subsection{Point Cloud Processing for DBH Estimation}

Extracting DBH from point clouds, whether from LiDAR or SfM, typically involves two main steps: segmenting individual tree stems and fitting geometric models to cross-sections.

\subsubsection{Tree Trunk Segmentation}
Various methods exist to isolate tree stem points from the surrounding environment (ground, canopy, understory). Common approaches include:
\begin{itemize}
    \item \textbf{Clustering algorithms:} Methods like DBSCAN and K-means group points based on spatial proximity or density, often applied to horizontal slices of the point cloud to identify potential stem locations \cite{ChenWangHangLi, f13040566}. 
    \item \textbf{Region Growing:} Starting from seed points (often identified in the trunk region), these algorithms iteratively add neighboring points based on criteria like proximity, normal vector similarity, or smoothness \cite{RegionGrowing}.
    \item \textbf{Projection-based methods:} These methods often project the 3D point cloud onto a 2D plane (e.g., XY plane) and use image processing techniques like Hough transform or density analysis to detect circular/cylindrical structures representing stems \cite{MICHALOWSKA2023100863}.
    \item \textbf{Deep Learning:} Recent approaches utilize deep neural networks (e.g., TreeLearn~\cite{HENRICH2024102888} and PointNet++~\cite{qi2017pointnet++, f14061159}) directly on the point cloud or its voxelized/projected representations for semantic or instance segmentation. Such deep learning-based methods require curating training data. We instead leverage large foundation vision models to get 2D semantic masks which are then projected onto 3D point clouds.
\end{itemize}

\subsubsection{Geometric Fitting for Diameter Estimation}
Once a stem is segmented, a cross-section at breast height (1.3m) is typically extracted, and a geometric shape is fitted to estimate the diameter. Common fitting methods include circle/cylinder, ellipse, B-spline, and Fourier \cite{7805274}. To handle noise and outliers common in point cloud data, robust estimation techniques like RANSAC (Random Sample Consensus) are often employed in conjunction with fitting algorithms.

\begin{figure}[!t]
\centering
\includegraphics[width=0.9\linewidth]{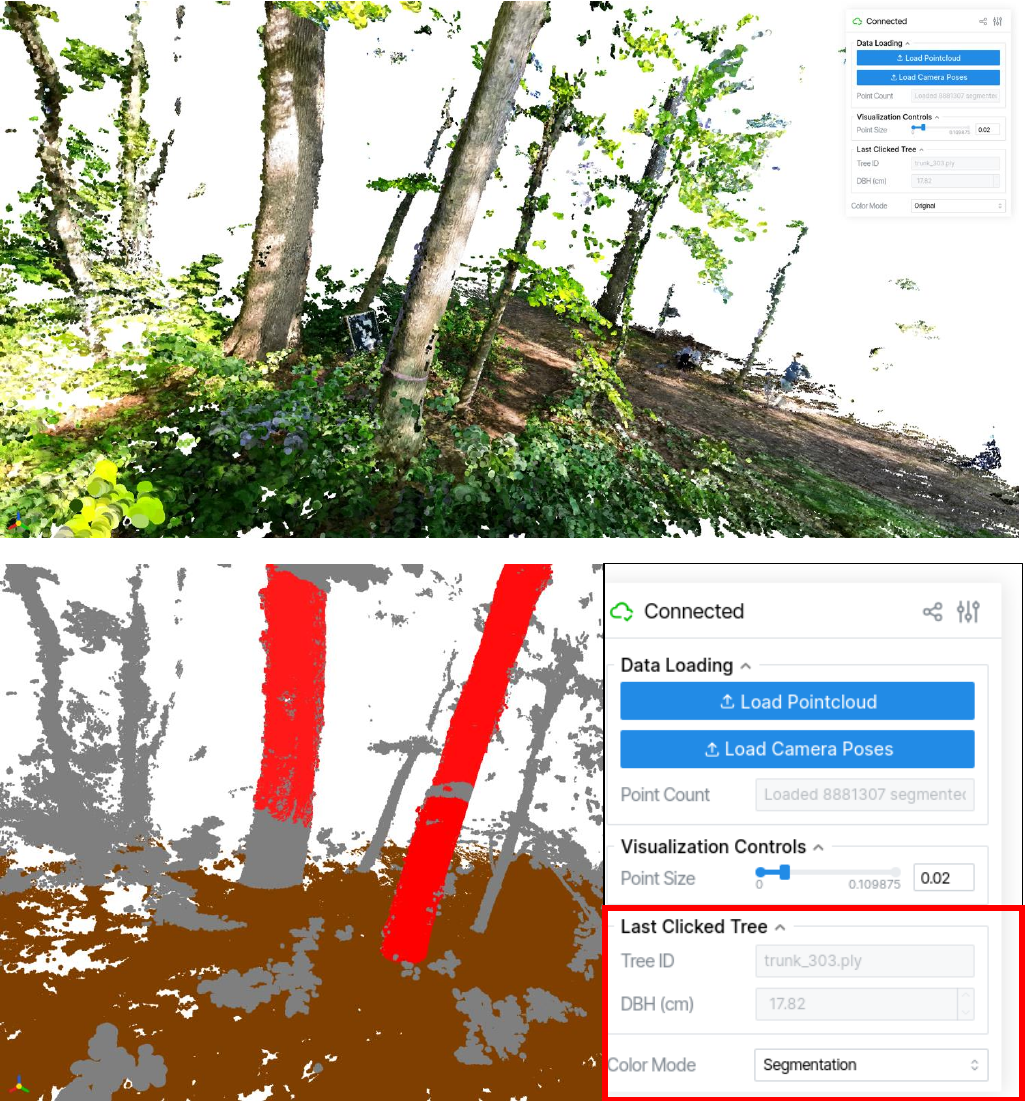} 
\caption{Screenshots of the custom visualization interface. Top: Photogrammetry point cloud rendered with original colors. Bottom left: Segmentation view highlighting identified tree trunks (in red). Bottom right: Sidebar panel showing tree ID and DBH estimate when a user clicks the tree trunk in the visualizer.}
\label{fig:visualizer}
\end{figure}

\section{Methodology}
\label{sec:methodology}

\subsection{System Overview}
The proposed system follows a semi-automated pipeline to estimate tree DBH from 360° video input. The overall workflow is depicted in Fig. \ref{fig:pipeline} and consists of the following main stages.
\begin{enumerate}
    \item \textbf{Data Acquisition:} Capturing 360° video footage of the target trees.
    \item \textbf{3D Reconstruction:} Generating a 3D point cloud using SfM-MVS from the video frames via Agisoft Metashape.
    \item \textbf{Scaling:} Manually scaling the reconstructed point cloud to real-world units.
    \item \textbf{Automated Tree Trunk Segmentation:} Isolating individual tree trunk point clouds using projected 2D semantic segmentations followed by extensive 3D post-processing.
    \item \textbf{Automated DBH Estimation:} Fitting geometric models (Ellipse, B-Spline, Fourier Series) to cross-sections at breast height using a RANSAC framework.
    \item \textbf{Visualization:} Displaying the segmented point clouds and estimated DBH values using a custom tool.
    \item \textbf{Evaluation:} Comparing the estimated DBH values against ground truth measurements.
\end{enumerate}

\subsection{Data Acquisition}
Data collection was conducted in Stadium Woods, located on the campus of Virginia Tech. A consumer-grade Insta360 omnidirectional camera and an Ouster LiDAR sensor were used to capture both visual and depth data for analysis. The LiDAR point cloud was collected to serve as a comparative baseline for evaluating the accuracy of DBH measurements produced by our SfM-based pipeline. Additionally, ground truth DBH measurements for selected trees were provided by Virginia Tech’s Department of Forest Resources and Environmental Conservation (FREC) and were used to quantitatively assess the performance of our method.

Two data acquisition platforms were employed. The first involved mounting the camera and LiDAR on a drone, which was flown in an orbit around individual trees to capture trunk geometry from various perspectives. The second platform consisted of a backpack-mounted tower system, allowing a human operator to walk around each tree to collect image and LiDAR data from ground level. In both approaches, care was taken to ensure sufficient coverage of the tree trunk from multiple viewpoints.

To enable accurate metric scaling of the reconstructed scenes, a 1 m × 1 m ArUco marker was placed in the field of view during each data collection session. This marker was later used to scale the 3D reconstruction to metric units.

\subsection{3D Reconstruction and Scaling}
We utilized the commercially available software Agisoft Metashape (formerly PhotoScan) to process the 360° video data. Metashape employs SfM to automatically estimate camera poses—both position and orientation—as well as intrinsic camera parameters by detecting and matching keypoints across overlapping images. This process, known as the alignment step, generates a sparse point cloud of the scene. A key advantage of Metashape for our application is its native support for equirectangular images, which is essential for processing imagery from omnidirectional cameras. Following alignment, MVS is applied to produce a dense 3D point cloud. We used standard processing settings within Metashape, including “High” quality for both alignment and dense cloud generation. From the dense point cloud, a mesh was generated and textured using the original images, resulting in a photorealistic 3D reconstruction.

As SfM inherently produces reconstructions in an arbitrary coordinate frame and scale, we manually scaled the resulting model using a 1 m × 1 m ArUco marker placed in the scene during data acquisition. In Metashape, we selected points on the marker with known distances in both the horizontal and vertical directions and used these references to compute a scaling factor, which was then applied to the entire point cloud and mesh. While this manual scaling approach was effective, it currently represents the primary barrier to a fully autonomous pipeline and introduces a potential source of error if reference measurements or point identification are inaccurate. Automating this step is a focus of ongoing work.

\begin{figure}[!t]
\centering
\includegraphics[width=0.45\linewidth]{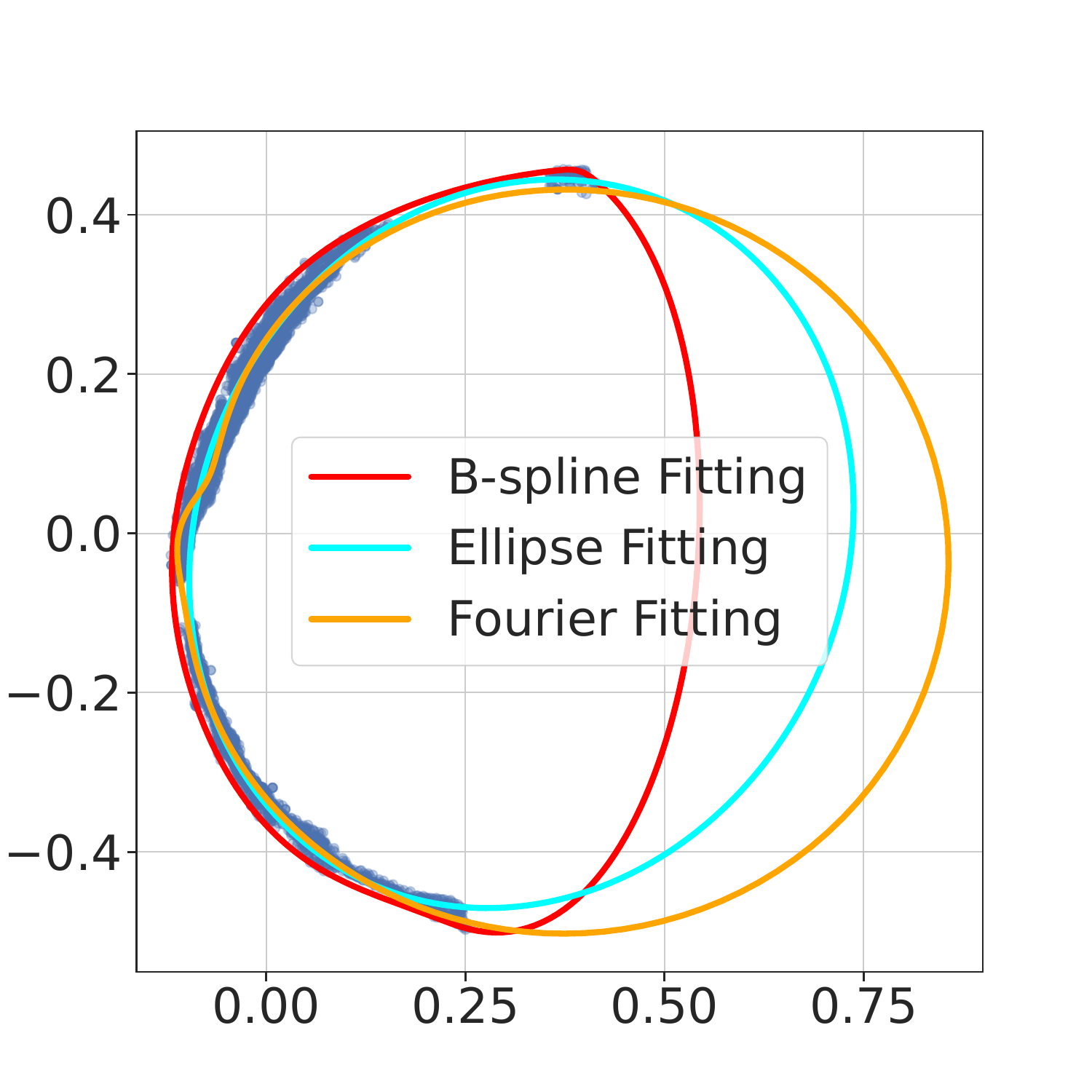}
\includegraphics[width=0.45\linewidth]{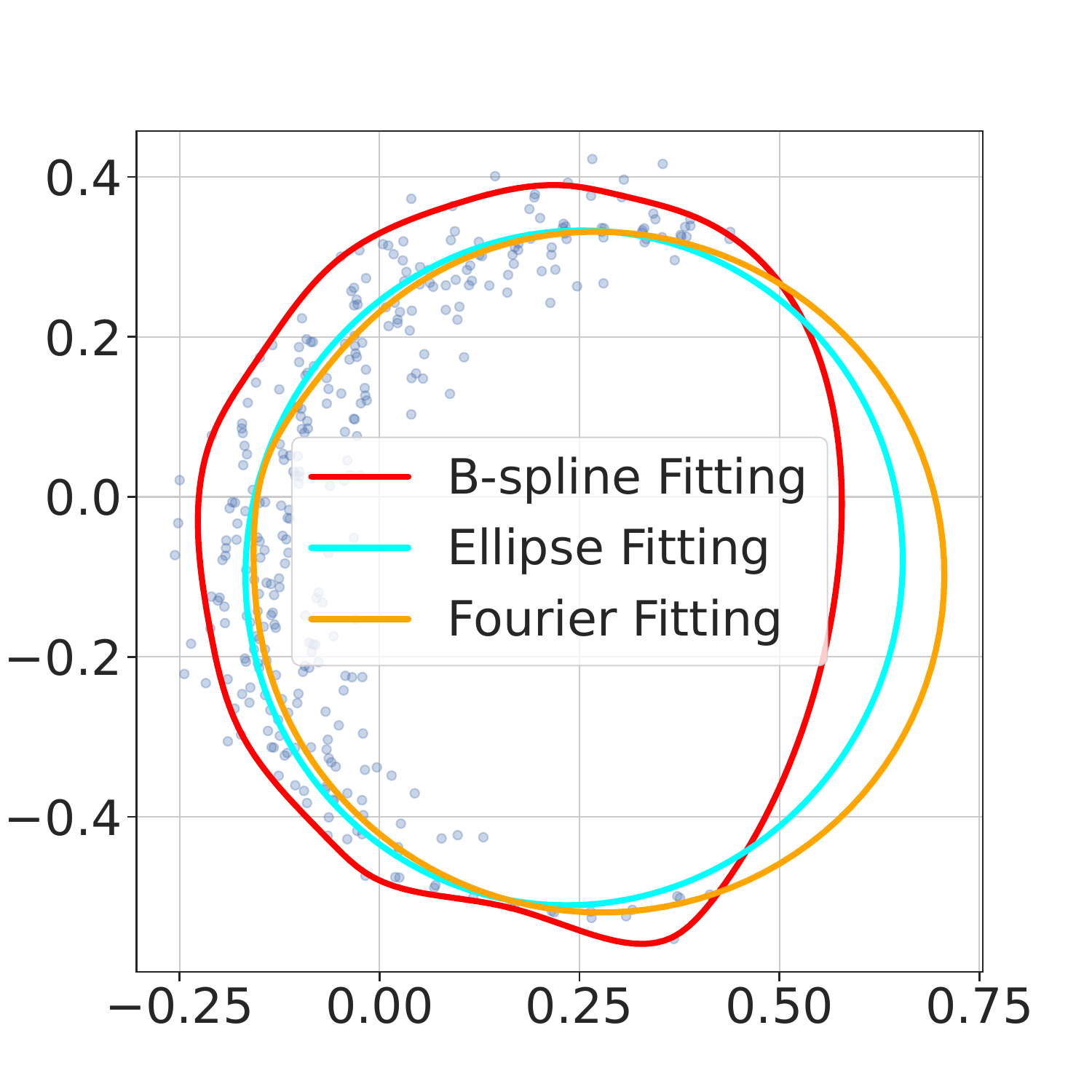}
\includegraphics[width=0.45\linewidth]{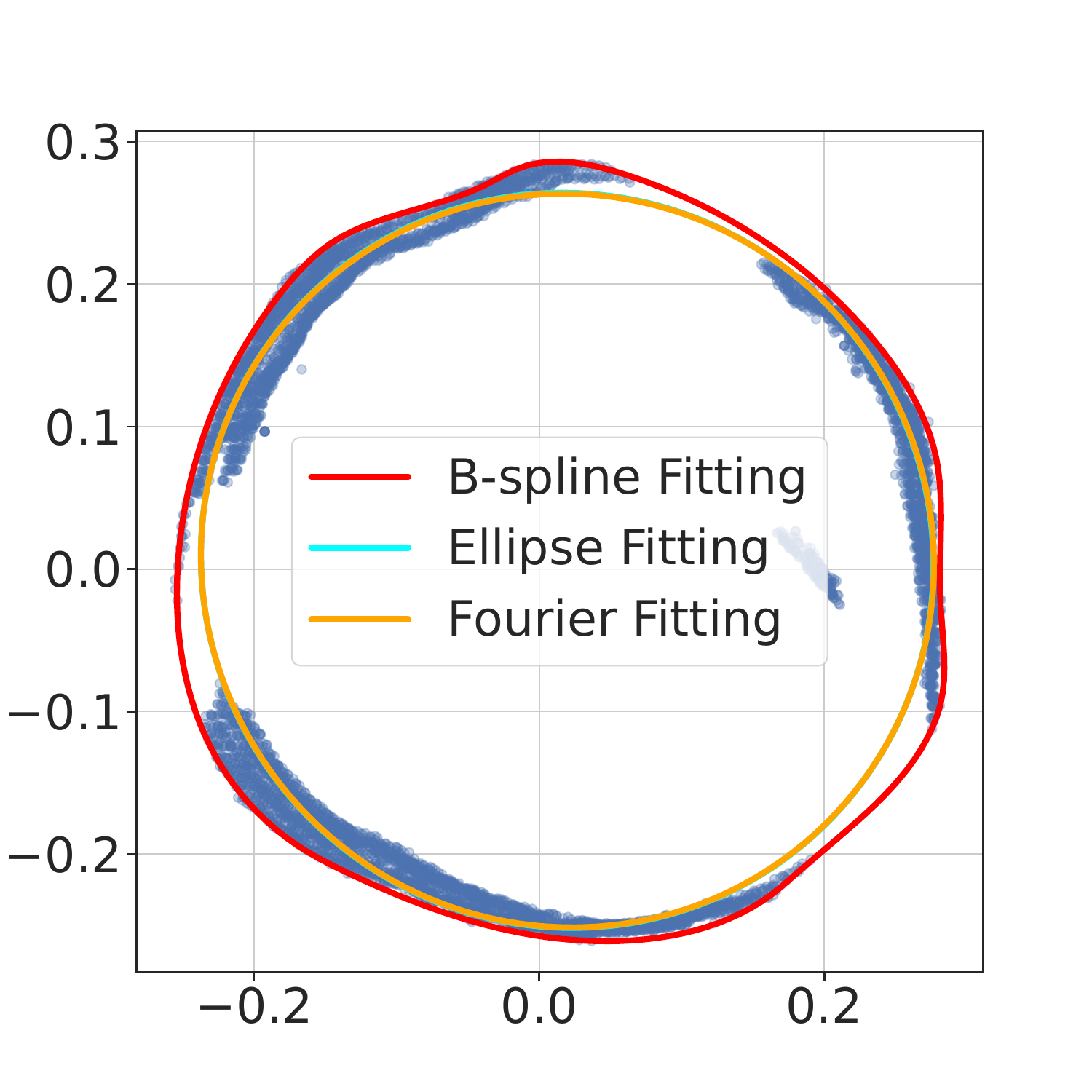}
\includegraphics[width=0.45\linewidth]{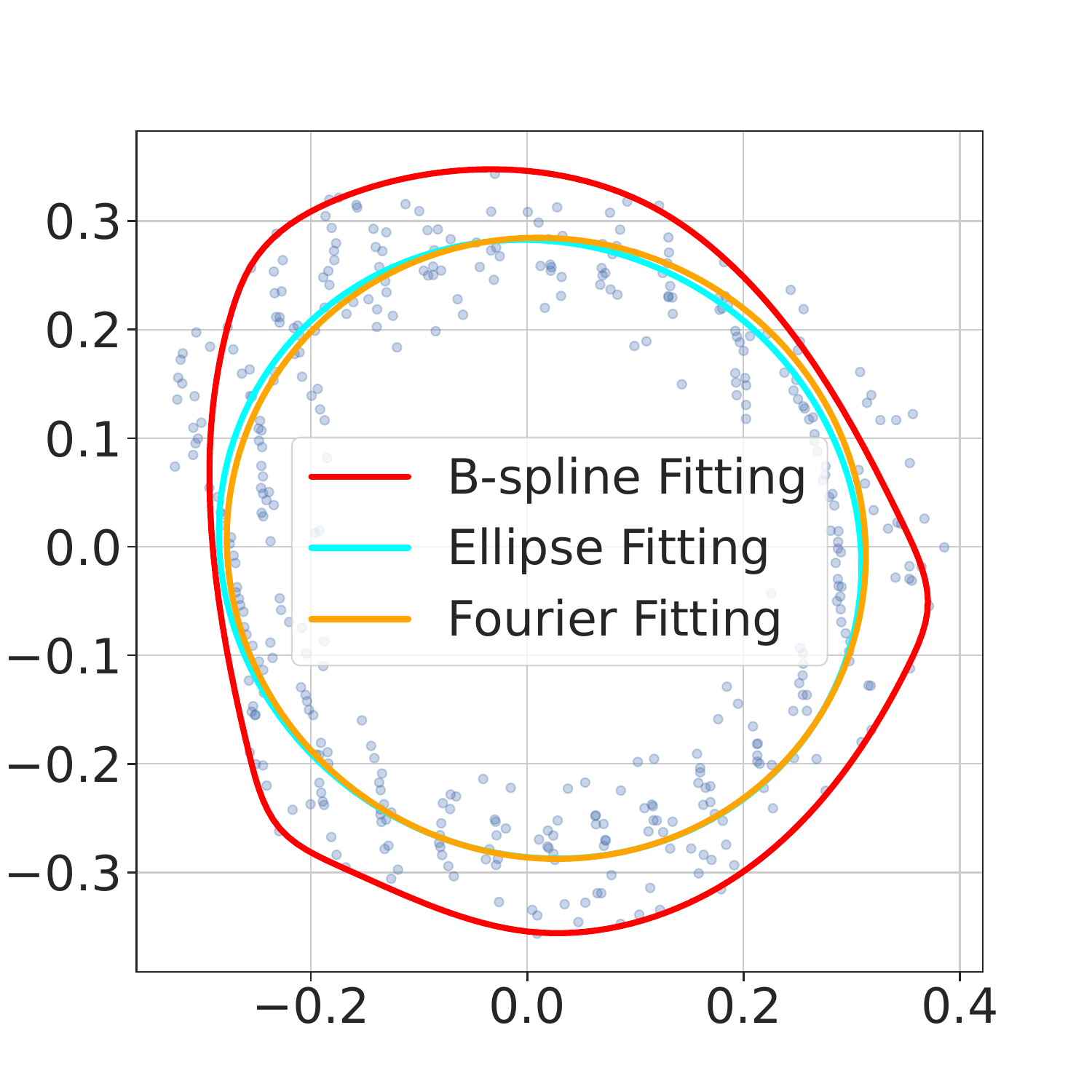}
\caption{Comparison of geometric fitting methods on cross-sections from both image-based (left column) and LiDAR-based (right column) point clouds. Top row: Partially observed trees. Bottom row: Fully observed trees. The truncated Fourier series (orange) provides more robust fits than ellipse (cyan) and B-spline (red) methods. Absolute relative errors for the Fourier method are: 6.18\% and 3.94\% (image-based), and 2.08\% and 10.69\% (LiDAR-based).}
\label{fig:crosssec}
\end{figure}

\subsection{Automated Tree Trunk Segmentation}

\subsubsection{Initial Segmentation via Projection}
We leverage Grounded SAM \cite{ren2024groundedsamassemblingopenworld}, a large vision foundation model, for initial segmentation of the trunks. Given the text prompt ``tree trunk'', the model produces segmentation masks identifying potential tree trunks in equirectangular images extracted from the 360° video. These 2D segmentation masks are then projected onto the 3D scaled point cloud. 

For each image $I$ with pose $\mathbf{T}=(\mathbf{R},\mathbf{t})$ and the set of $N$ points in the scaled point cloud  $\mathcal{P}=\{\mathbf{p}_i\}_{i=1}^N\subset\mathbb{R}^3$, we project every point $\mathbf{p}_i$ into the spherical image via
\[
\mathbf{d}_{i} \;=\; \frac{\mathbf{R}^\top(\mathbf{p}_i-\mathbf{t})}{\|\mathbf{p}_i-\mathbf{t}\|}, 
\qquad
(u_{ij},v_{ij})
=\mathrm{EQR}(\mathbf{d}_{ij};\,H,W),
\]
where $\mathrm{EQR}(\cdot)$ denotes the equirectangular mapping to pixel coordinates in an image of height $H$ and width $W$. We then accumulate a \emph{label count} for each point. If a point is projected into an image pixel classified as tree trunk, we increment the point's \emph{label count} by 1. Points with \emph{label count} of at least one form the initial set of \emph{trunk candidates}.

\subsubsection{Ground Plane Removal and Normal Filtering}
To eliminate spurious trunk candidates, we first estimate the dominant ground plane via RANSAC on the full point cloud. We discard all points near the ground plane. Next, we compute normals on the remaining cloud and remove points that are not sufficiently parallel to the ground. This filtering is based on the implicit assumption that trees do not tilt by more than 45 degrees.

\subsubsection{Cluster Extraction and Expansion}
From the filtered trunk candidates, we perform DBSCAN clustering \cite{10.5555/3001460.3001507} directly on the remaining points. To remove noise (clusters that are not tree trunks), we compute each cluster’s size, apply Otsu’s method \cite{4310076} to find the threshold, and discard clusters that are below the threshold. To further recover trunk points missed by the initial projection and filtering, we expand each retained cluster. Specifically, points that are close to the points in the cluster are merged into the cluster.

\subsubsection{Transferring Trunk Segmentations to the LiDAR Point Cloud}
To compare the photogrammetry‑based DBH estimation with LiDAR‑inertial odometry, we transfer these point‑cloud labels from the photogrammetric model to the LiDAR cloud.  We first align the two modalities by fitting a Procrustes transform to their respective sensor trajectories and applying it to the LiDAR scan.  Finally, we build a spatial index (cKDTree) on the transformed LiDAR points and, for each photogrammetric trunk segment, collect all LiDAR points within a small radius (0.2 m).  These points inherit the trunk label and are written out as a corresponding LiDAR‐based trunk segment for direct quantitative and visual comparison.


\subsection{Estimating the Diameter at Breast Height}

Given the segmented trunk point cloud, we estimate the diameter at breast height (DBH) using three curve‐fitting methods on the cross‐section. All approaches operate on the 30 cm tall band between 1.22 m and 1.52 m above the estimated ground plane.

\subsubsection*{RANSAC‐based boundary fitting}
We project the cropped band onto the plane orthogonal to the trunk axis to obtain a planar cross‐section as shown in \cref{fig:crosssec}. A RANSAC scheme is used to robustly fit each of three closed‐curve models:
\begin{itemize}
  \item \textbf{Least‐squares ellipse fitting \cite{765658}.} 

  \item \textbf{Periodic B‐spline fitting \cite{7805274}.}  
    An initial boundary is extracted via an $\alpha$‐shape), then a periodic cubic B‐spline is fitted by minimizing the sum of squared distances between the spline curve and the boundary points under periodicity constraints. This yields a smooth, flexible representation that can capture fine‐scale irregularities.
  \item \textbf{Truncated Fourier series (degree 2) \cite{7805274}.}  
    We represent the closed curve parametrically in polar coordinates as
    \begin{align*}
      r(t) &= a_{0} \;+\;\sum_{i=1}^{2} a_{i}\cos(i\,t)+b_{i}\sin(i\,t)
    \end{align*}
    for \(t\in[0,2\pi)\).  The coefficients \(\{a_{0},a_{i},b_{i}\}_{i=1}^{2}\) are obtained by solving a single linear least‐squares problem over all cross‐section points.
\end{itemize}

\begin{figure*}[!t]
\centering

\subfloat[Full (left) vs. Partial (right) observability. Partial observations lead to greater variance and underestimation across all fitting methods. \Cref{fig:crosssec} shows the comparison of fully and partially observed cross sections. \label{fig:fullvspartial}]{
\begin{minipage}[t]{0.45\textwidth}
\centering
\includegraphics[width=0.48\linewidth]{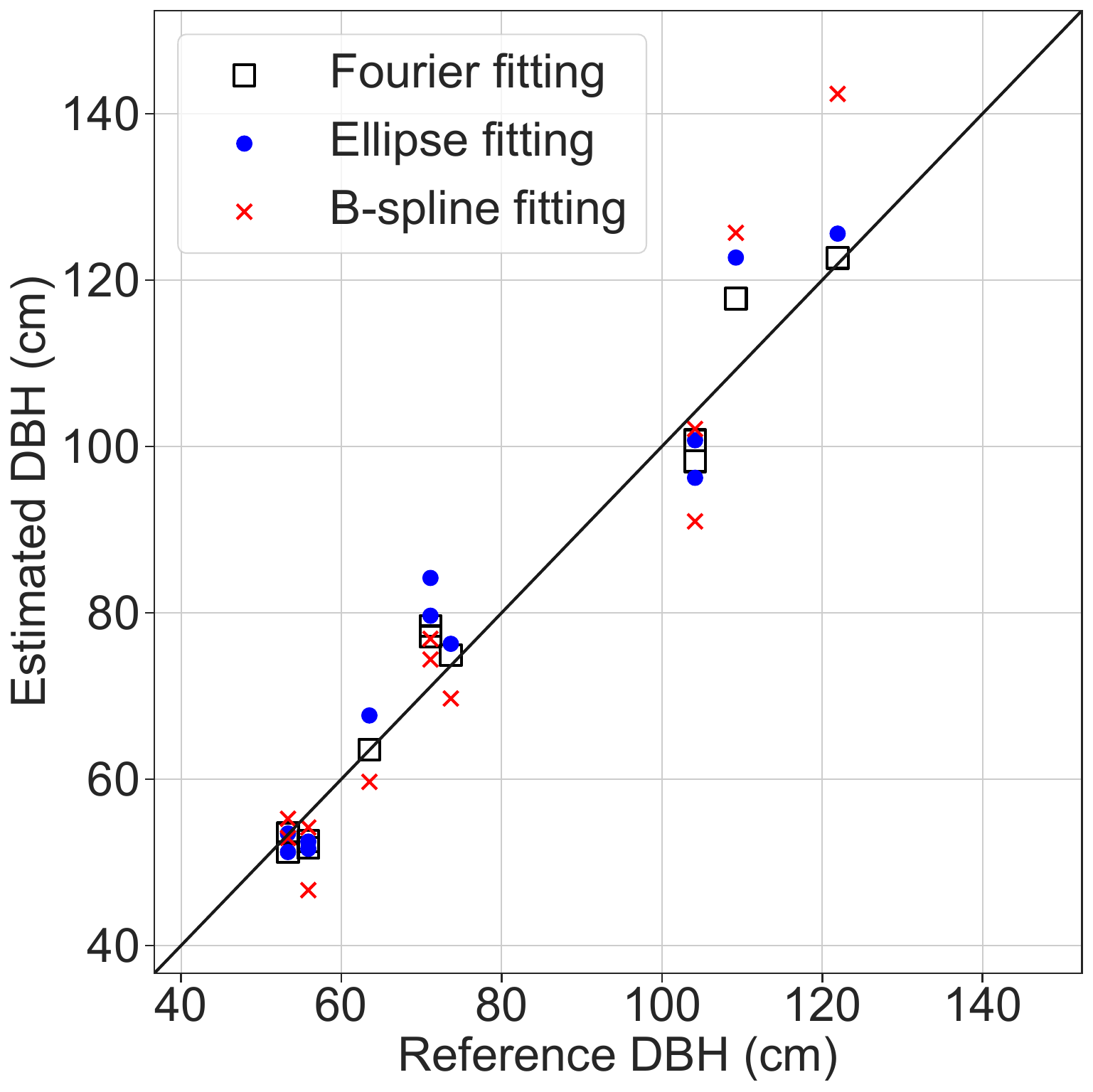}
\includegraphics[width=0.48\linewidth]{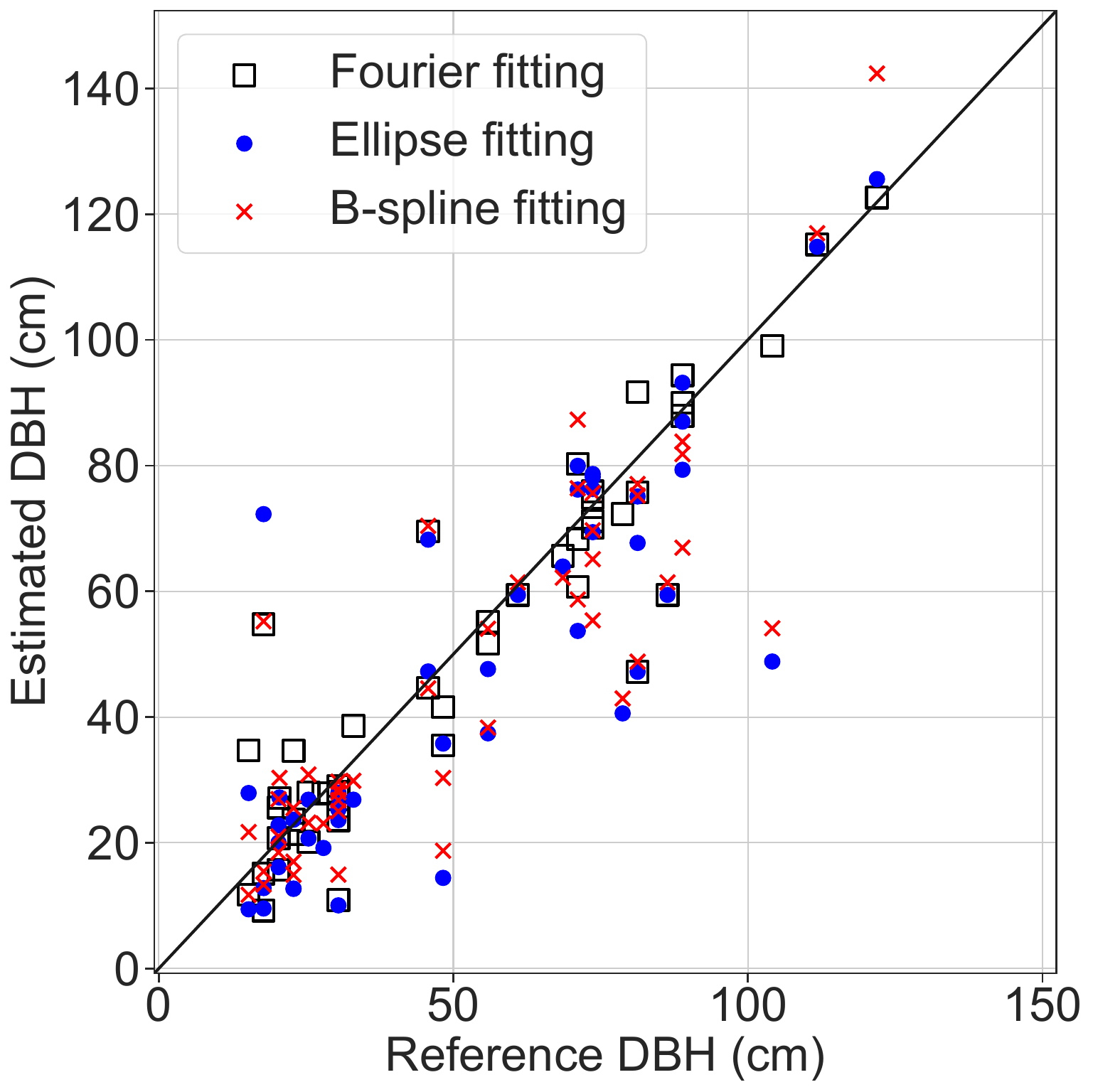}\\
\includegraphics[width=0.48\linewidth]{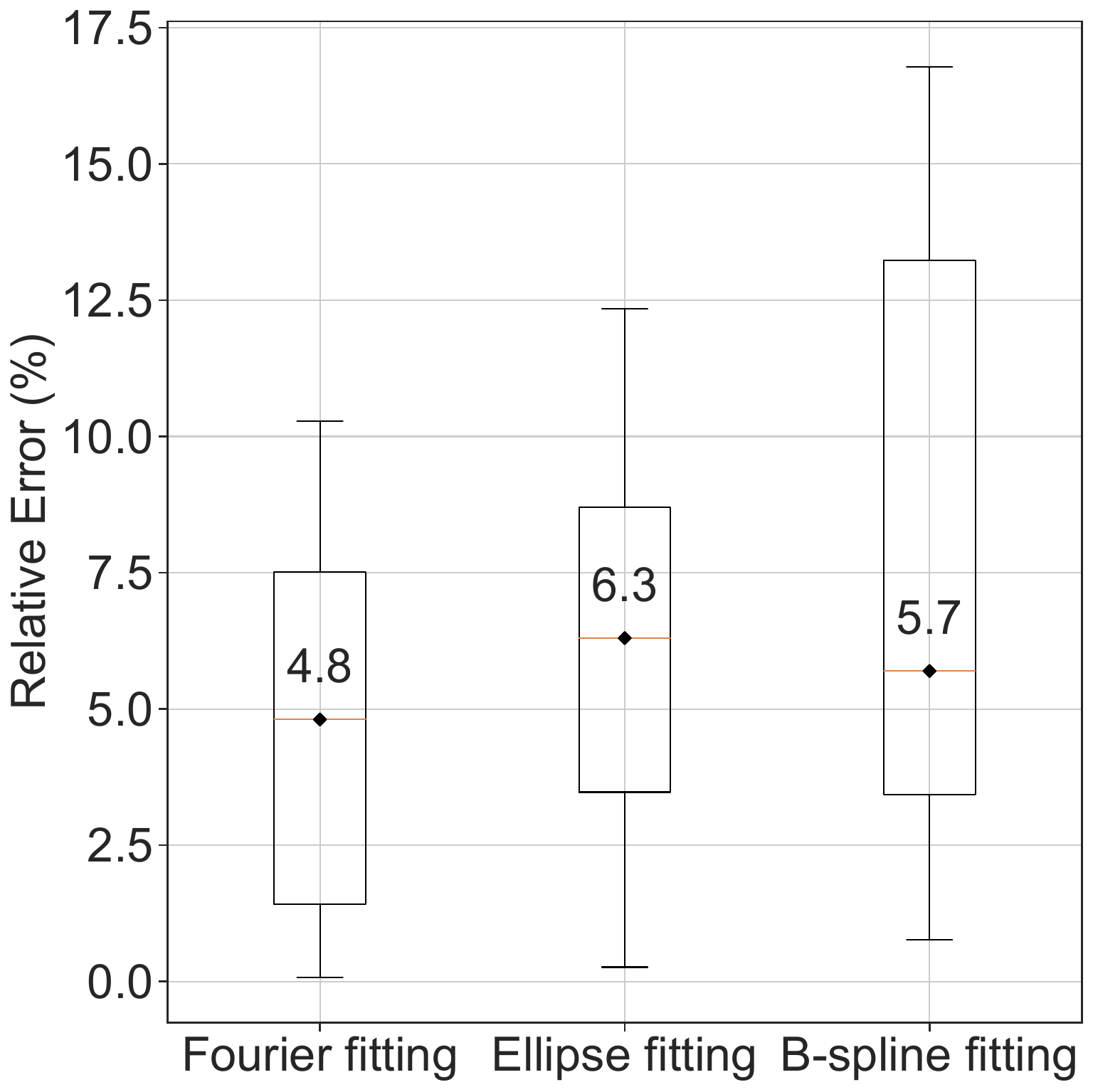}
\includegraphics[width=0.48\linewidth]{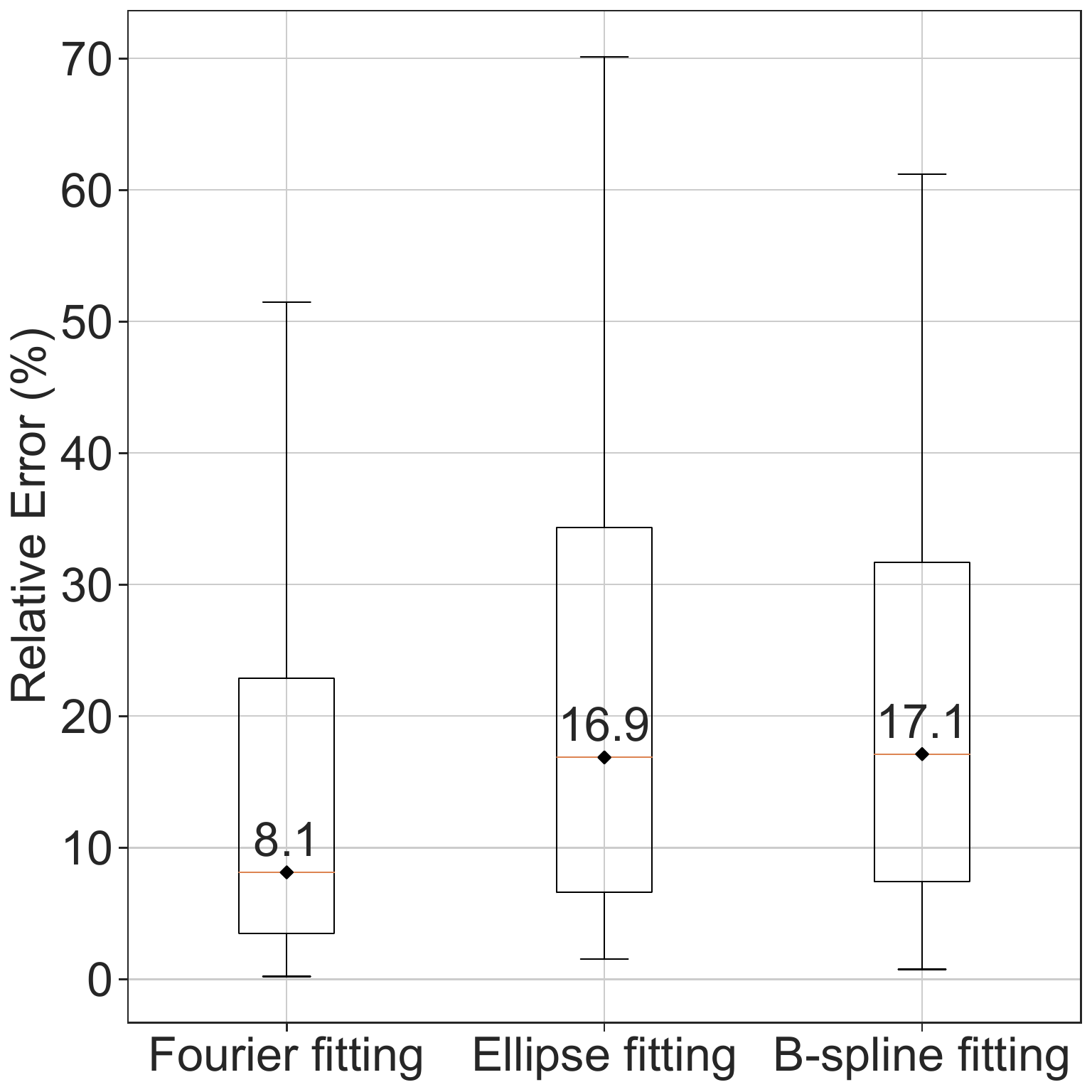}
\end{minipage}
}
\hfill
\subfloat[Backpack (left) vs. Drone (right) platforms. Fourier fitting consistently yields the lowest error, while drone data suffers from motion blur and incomplete coverage. The two platforms are shown in \cref{fig:pipeline}. \label{fig:backpackvsdrone}]{
\begin{minipage}[t]{0.45\textwidth}
\centering
\includegraphics[width=0.48\linewidth]{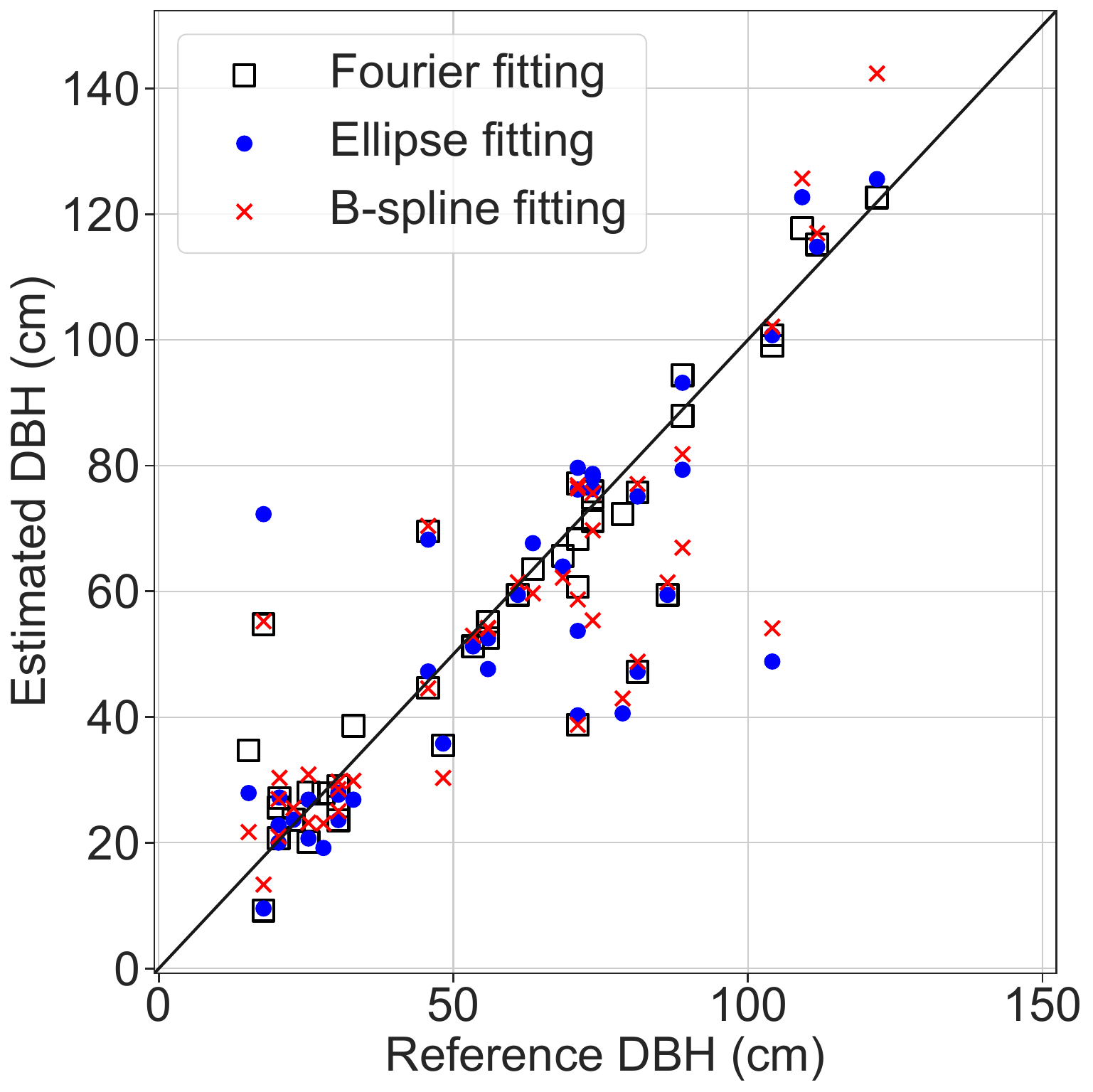}
\includegraphics[width=0.48\linewidth]{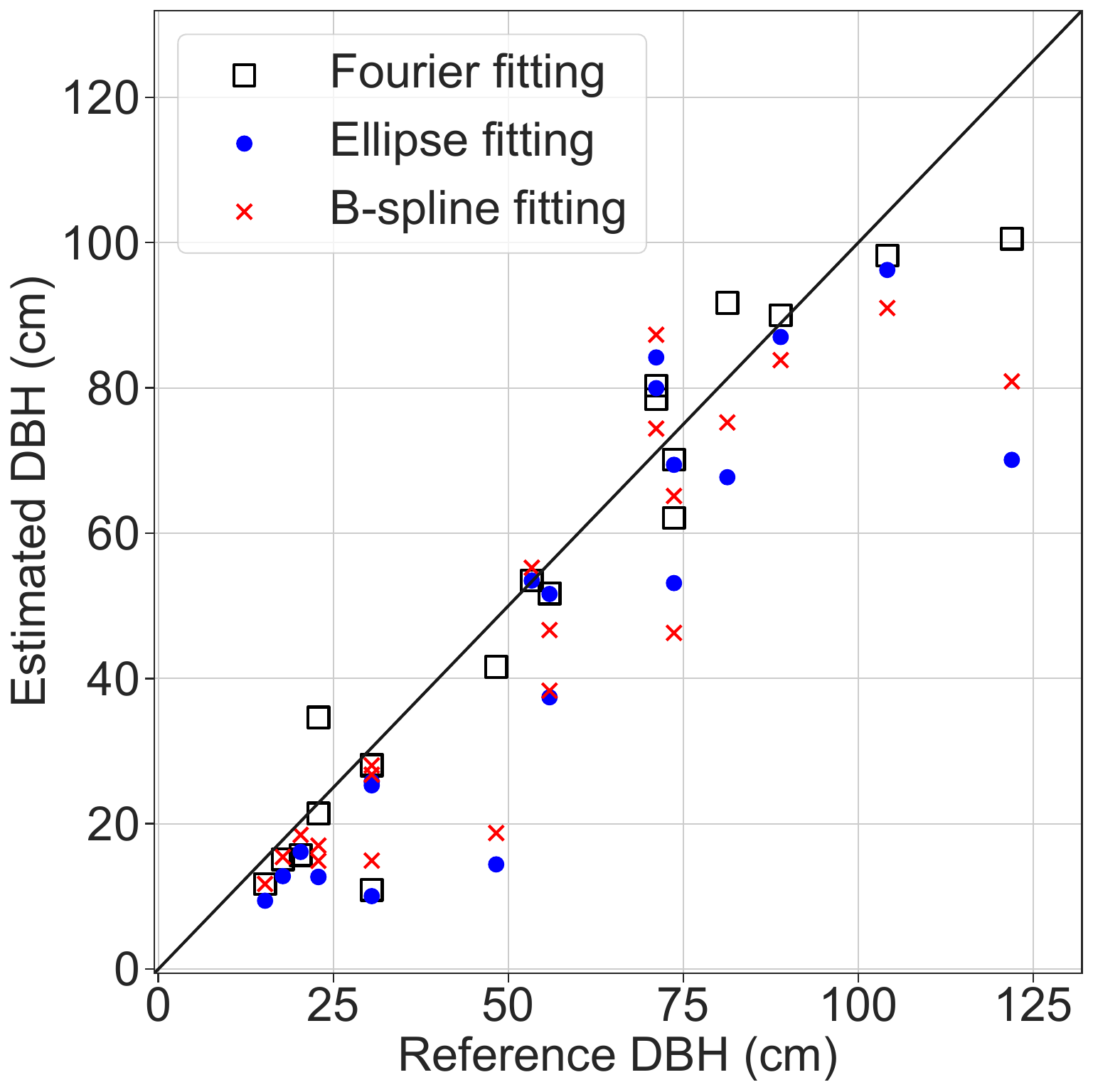}\\
\includegraphics[width=0.48\linewidth]{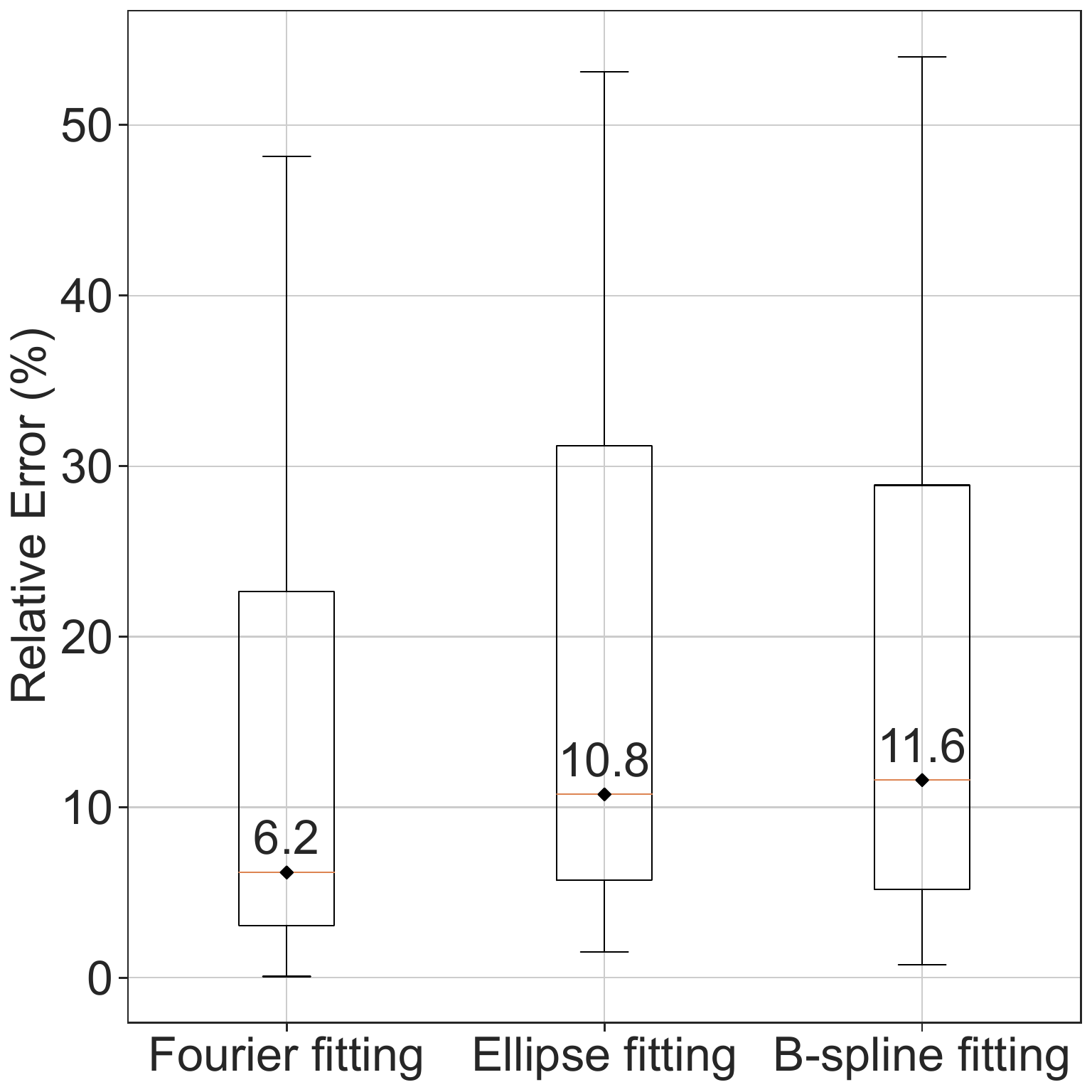}
\includegraphics[width=0.48\linewidth]{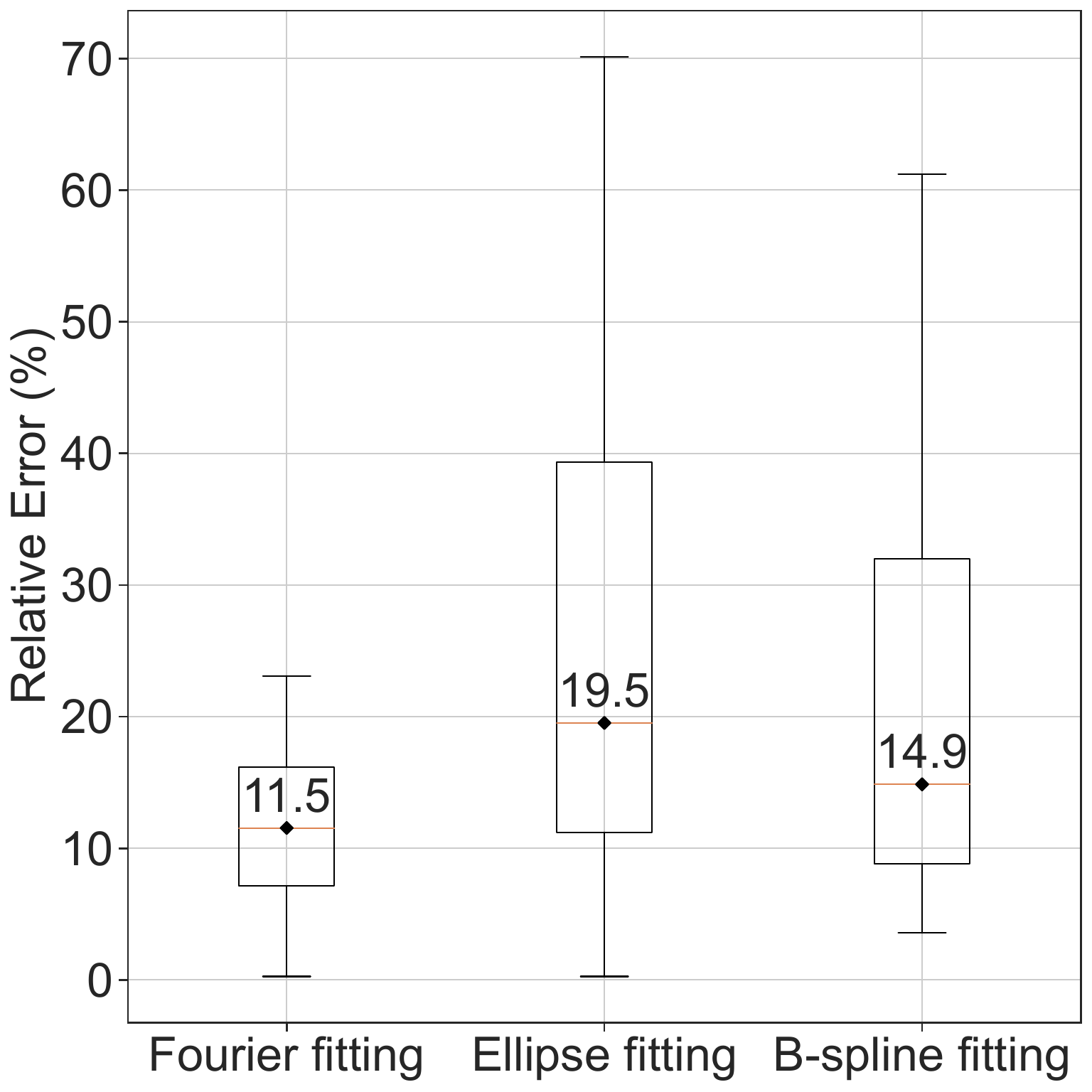}
\end{minipage}
}


\subfloat[Short (left) vs. Long (right) Sequences. Short trajectories yield more accurate DBH estimates due to better cross-section visibility. Long trajectories incur higher error due to drift and sparse sampling. Fourier fitting remains the most robust across sequence lengths. Examples of short and long sequences are shown in ~\cref{fig:longshortreconstruction} \label{fig:longvsshort}]{
\begin{minipage}[t]{0.45\textwidth}
\centering
\includegraphics[width=0.48\linewidth]{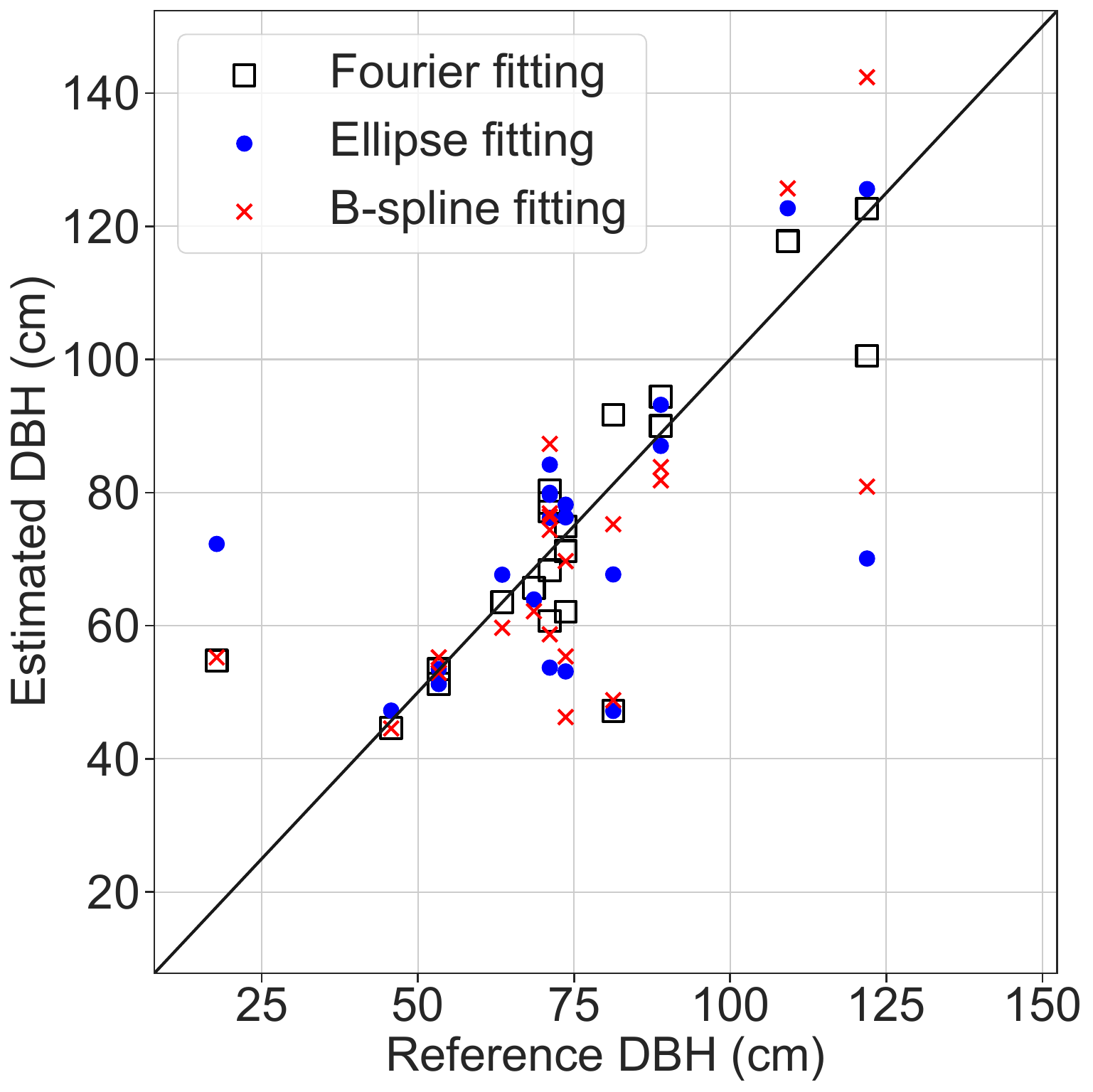}
\includegraphics[width=0.48\linewidth]{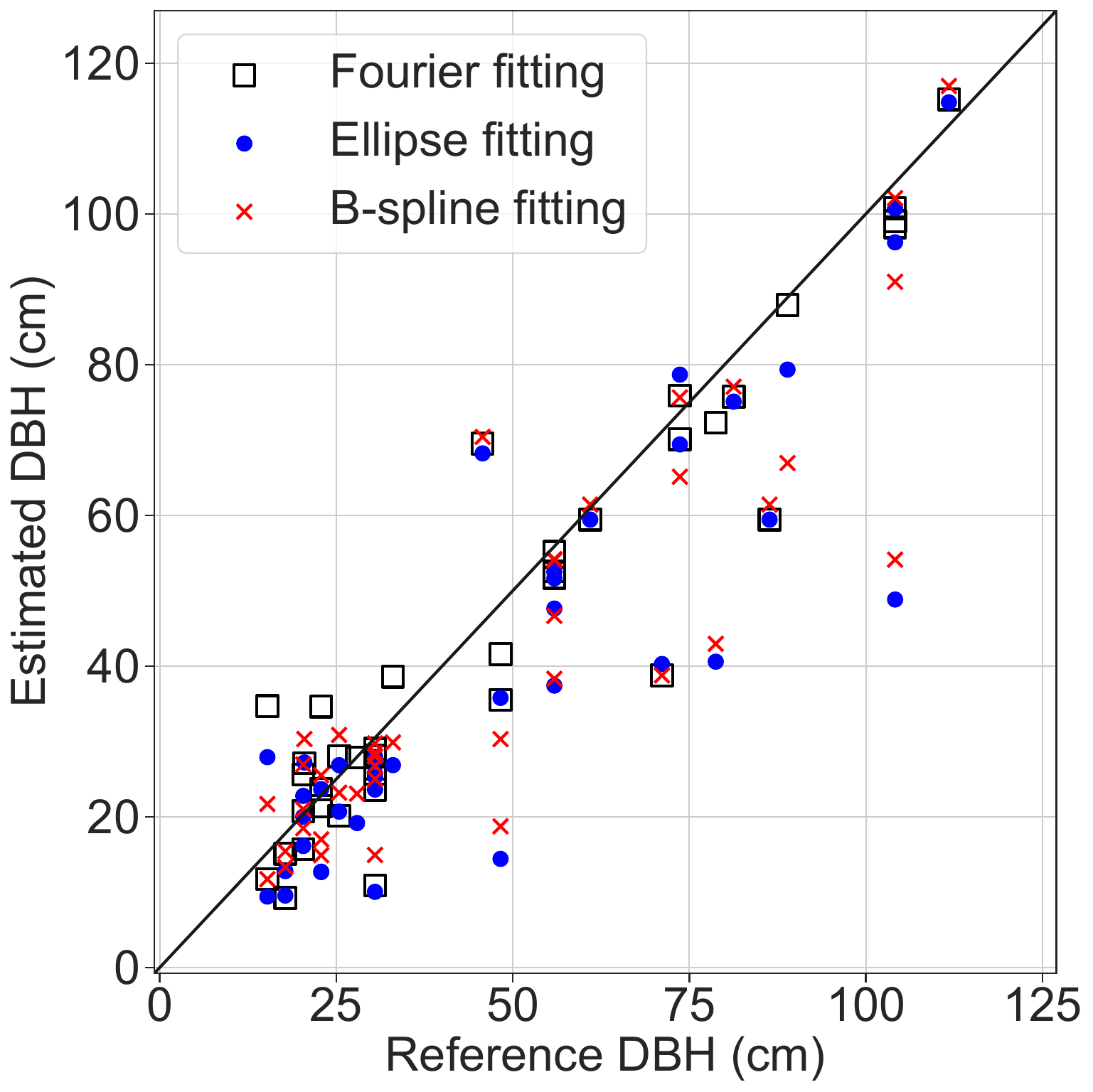}\\
\includegraphics[width=0.48\linewidth]{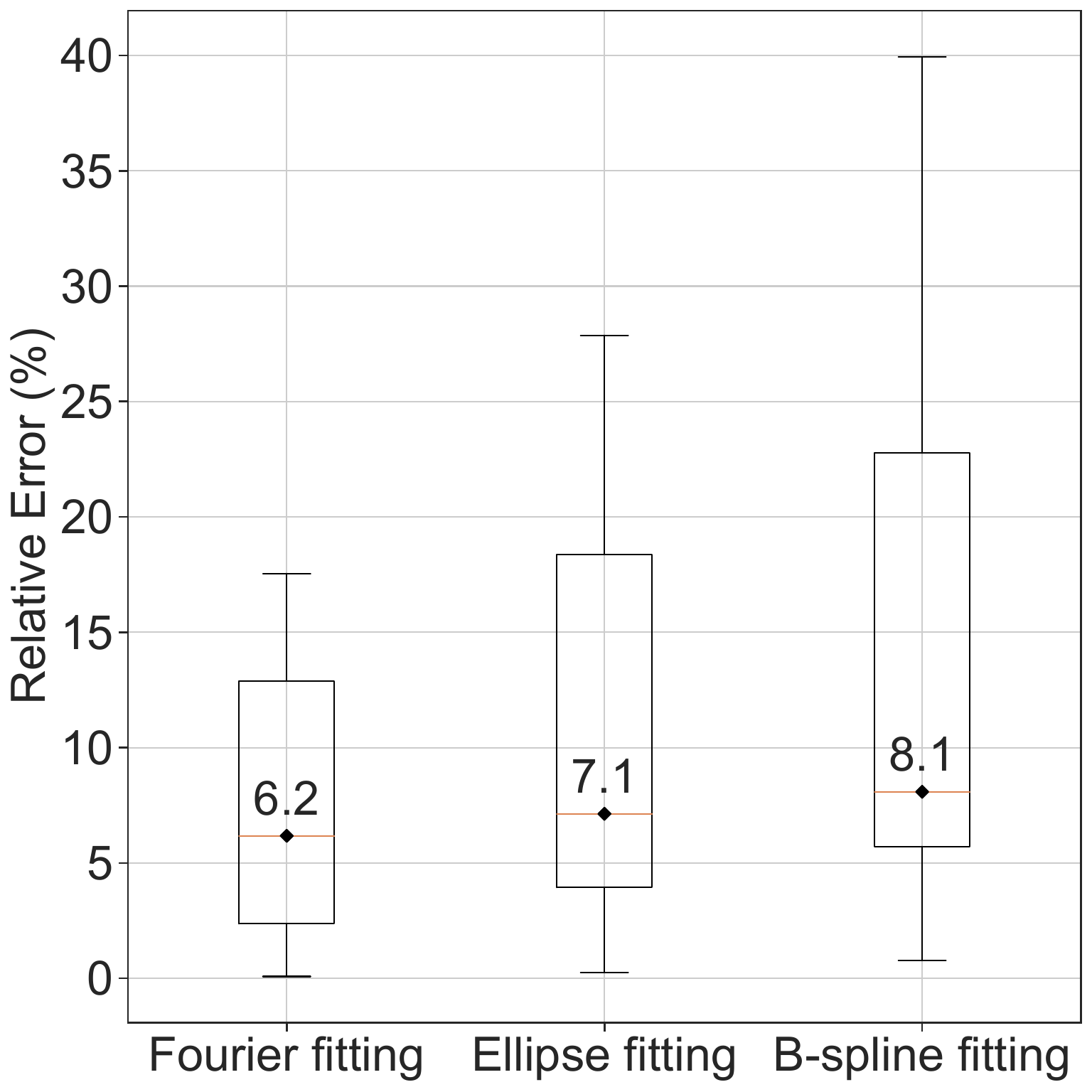}
\includegraphics[width=0.48\linewidth]{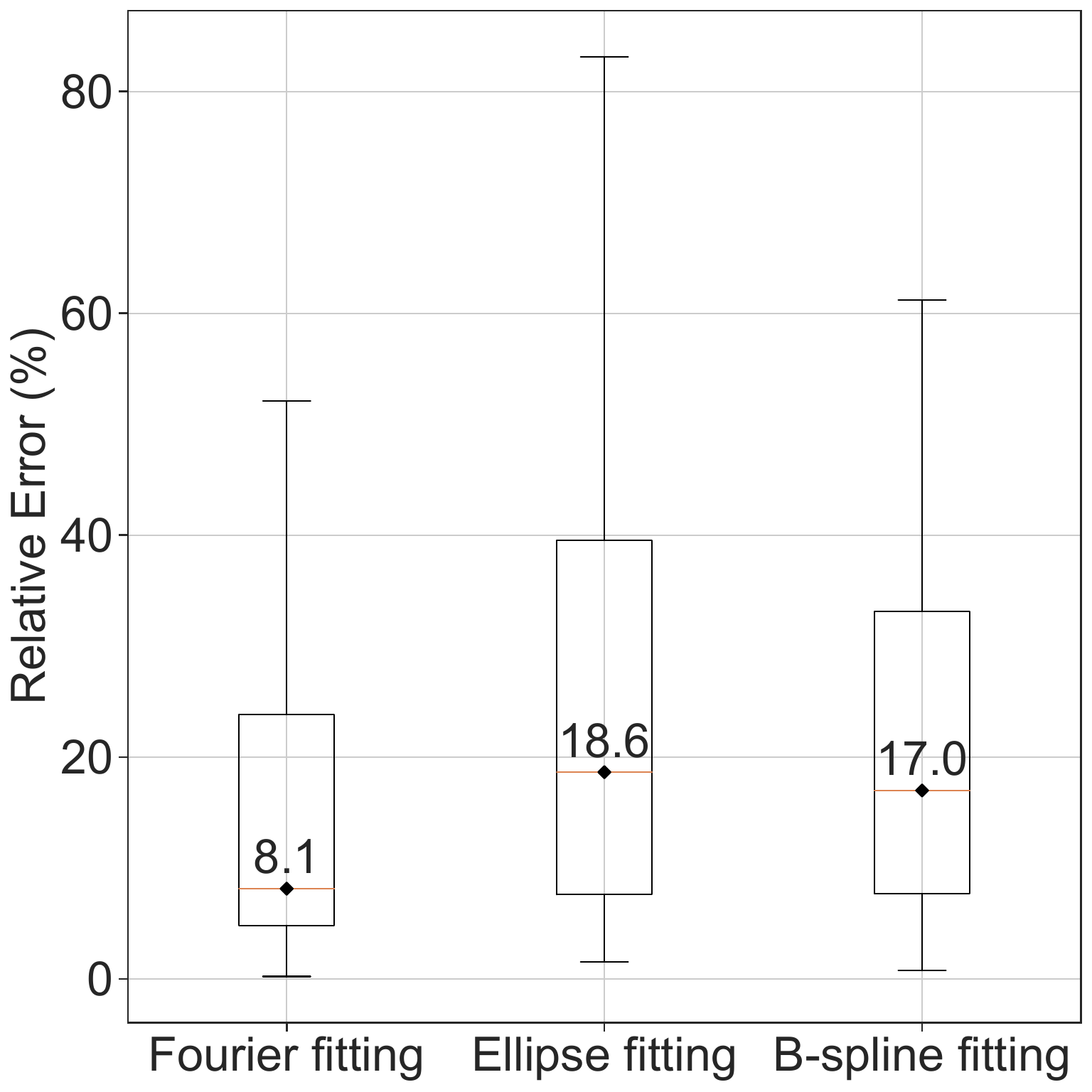}
\end{minipage}
}
\hfill
\subfloat[Photogrammetry (left) vs. LiDAR-inertial odometry (right). Fourier fitting shows the best performance in both modalities. DBH estimation based on Photogrammetry outperforms DBH estimation based on LiDAR-inertial odometry. Cross-section fitting for both modality are compared in \cref{fig:crosssec}.  
\label{fig:imagevslidar}]{
\begin{minipage}[t]{0.45\textwidth}
\centering
\includegraphics[width=0.48\linewidth]{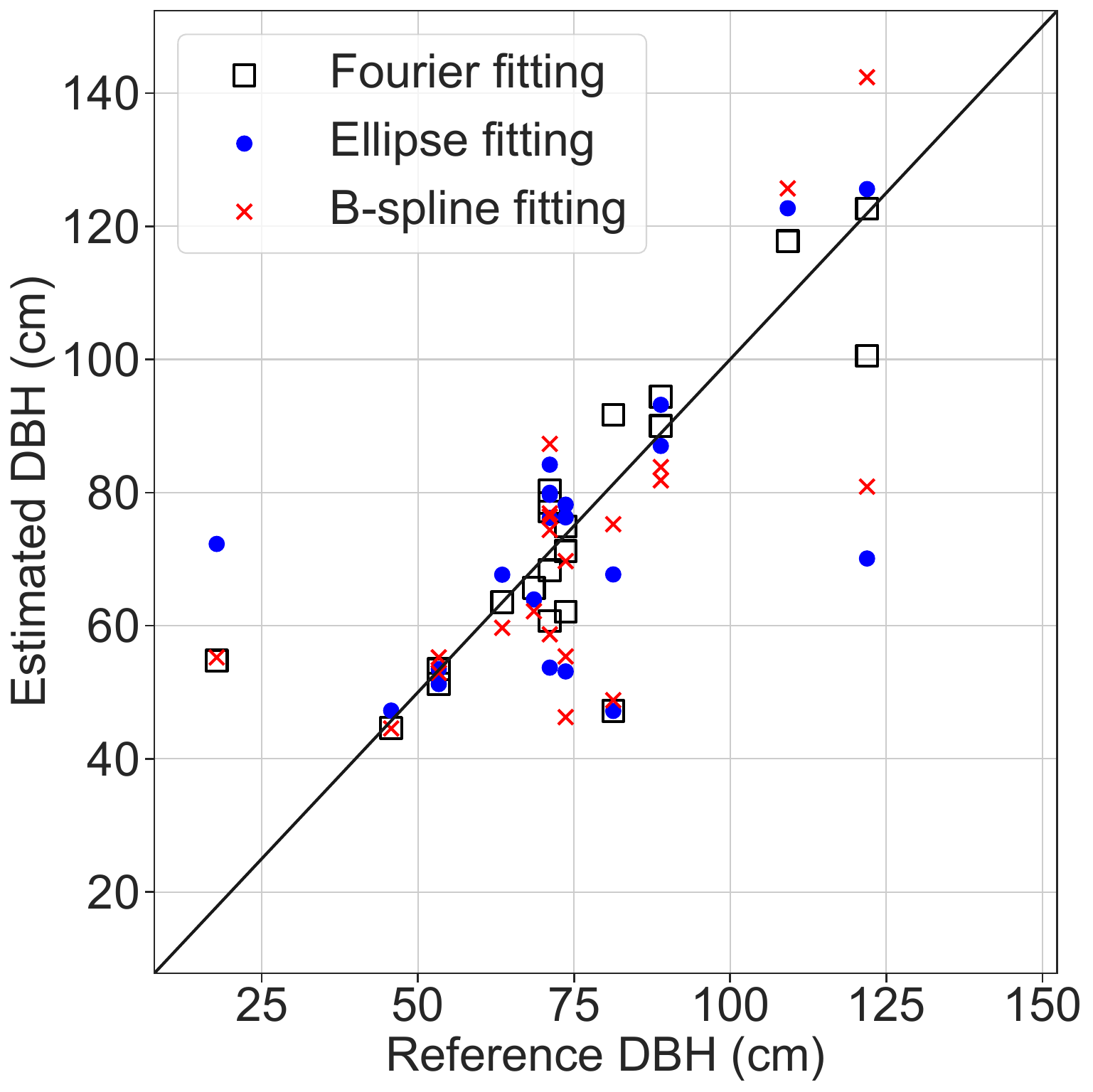}
\includegraphics[width=0.48\linewidth]{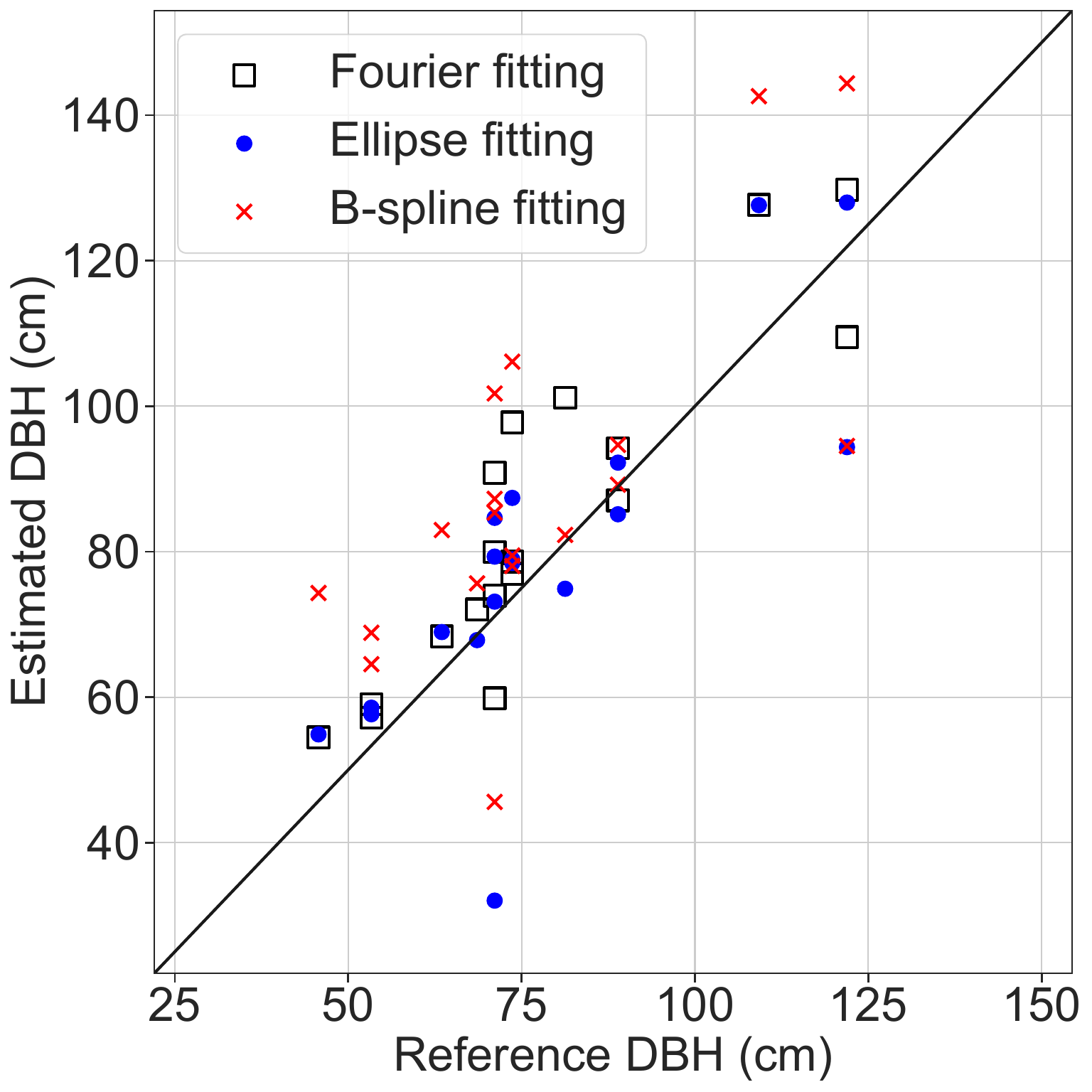}\\
\includegraphics[width=0.48\linewidth]{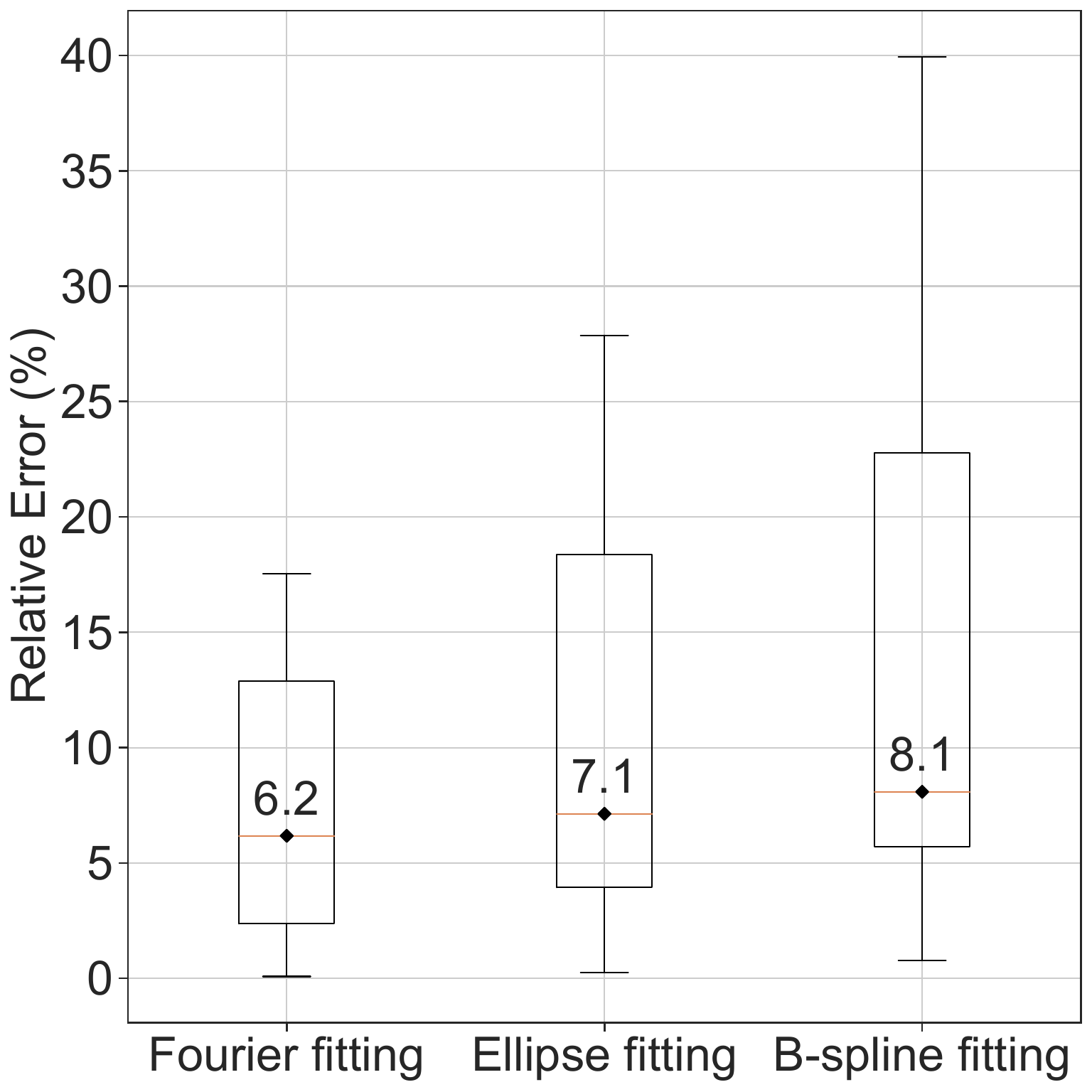}
\includegraphics[width=0.48\linewidth]{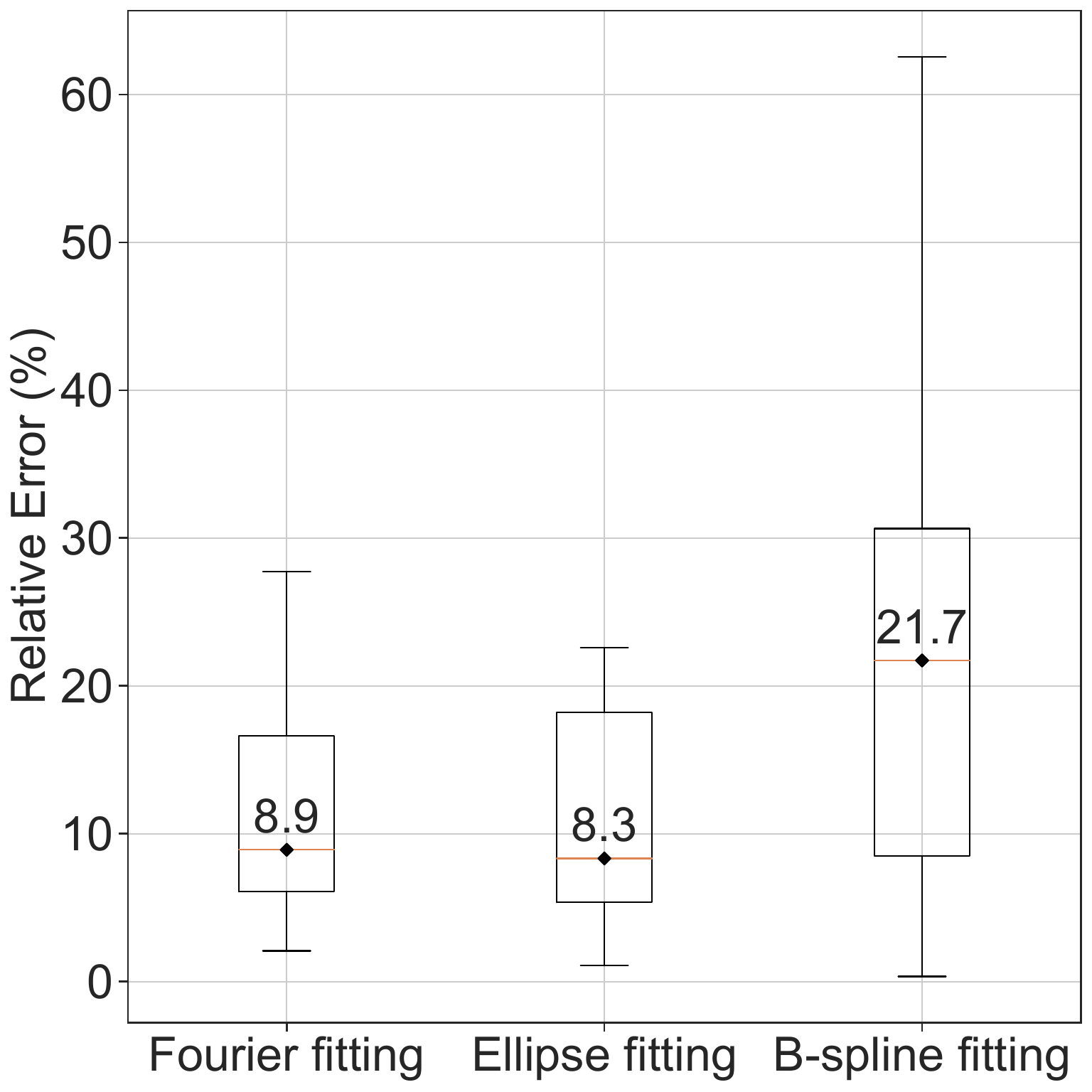}
\end{minipage}
}

\caption{Comparison of DBH estimation performance across observability, platform, sequence length, and sensing modality. Each block contains top-row Estimated vs. reference DBH scatter plots and bottom-row box plots of relative error (\%), comparing three fitting methods.}
\label{fig:biggrid}
\end{figure*}

In each RANSAC iteration, we randomly sample 50 points, fit the chosen curve, collect all inliers within 5 cm, and refit to those inliers. After 100 iterations, the model with the greatest inlier support is selected. The final perimeter of the fitted curve is divided by $\pi$ to produce the DBH estimate. Across all collection scenarios, the Fourier‐series method consistently yields the lowest median error and lowest variance compared to the other two methods as in \cref{fig:biggrid}.

\subsection{Visualization Tool}
To facilitate qualitative inspection and quantitative validation, we developed an interactive 3D visualizer based on viser \cite{viser}.  This tool allows users to load the processed point cloud data and interactively view the segmented trees (in ``Segmentation'' Color Mode, tree trunks are in red; in ``Original'' Color Mode, points have original color from photogrammetry).   By clicking on any trunk cluster, the tool displays its unique identifier and the corresponding DBH estimate in a sidebar (see \cref{fig:visualizer}).  

\section{Experiments and Results}
\label{sec:experiments}

\subsection{Experimental Setup}

\subsubsection{Dataset}
The proposed system was evaluated on a dataset comprising 61 samples of 43 trees located in 
Stadium Woods, Virginia Tech. As described in~\cref{tab:data-summary}, the trees exhibited variability in DBH. The trees also cover several species including Quercus alba (White Oak), Quercus velutina (Black Oak), Prunus avium  (Sweet Cherry), Prunus serotina  (Black Cherry), Acer saccharum (Sugar Maple), Acer rubrum (Red Maple), and Acer platanoides (Norway Maple). 360° video and LiDAR-inertial odometry estimates were obtained for each tree following the procedure outlined in Section \ref{sec:methodology}.

\subsubsection{Ground Truth}
Ground truth DBH for each of the 43 trees was measured manually using a standard diameter tape at 1.3 m above ground level \cite{VirginiaTech}.

\subsubsection{Comparison to LiDAR-based system}
To benchmark the performance of our low‑cost 360° camera pipeline, we compared its DBH estimates against those derived from a LiDAR‐based mapping system. We used Faster-LIO to build a dense LiDAR point cloud using a LiDAR following the same trajectory as the 360° camera. We also applied the identical DBH estimation pipeline to the LiDAR cloud. Although we did not re-implement the single‐scan cylinder‐fit approach used in prior work, hand-held and UAV‐mounted LiDAR systems have been shown to achieve a median absolute / relative error of 0.65 cm / 2.63\% and 1.45 cm / 5.59\% over a 10–60 cm DBH range in similar forest conditions \cite{PRABHU2024111050}. The results in our paper and \cite{PRABHU2024111050} are comparable since we use the same LiDAR and mobile platform, and evaluate on trees of similar DBH range in similar forests.

\subsection{Evaluation Metrics}

To quantify the accuracy and precision of our DBH estimates, we report four robust statistics for each fitting method and acquisition scenario. We choose to use robust statistics since there are bad estimates corresponding to highly occluded and far-away trees.

\begin{itemize}
  \item \textbf{Median bias.} Let $e_i = \hat y_i - y_i$ be the error between the estimated DBH $\hat y_i$ and the reference DBH $y_i$.  The median bias is 
  \[
    \mathrm{bias}_{\mathrm{med}} = \mathrm{median}\{\,e_i\},
  \]
  which measures systematic over‑ or under‑estimation.

  \item \textbf{Median absolute deviation (MAD).}  We compute the absolute deviations from the median error,
  \[
    d_i = \bigl|\,e_i - \mathrm{median}\{e_j\}\bigr|,
  \]
  and define
  \[
    \mathrm{MAD} = \mathrm{median}\{\,d_i\}.
  \]
  MAD summarizes typical scatter while rejecting outliers.

  \item \textbf{Robust coefficient of variation (rCV).}  To express dispersion relative to tree size, we divide MAD by the median reference DBH:
  \[
    \mathrm{rCV} = \frac{\mathrm{MAD}}{\mathrm{median}\{y_i\}}\times 100\%.
  \]
  Values between 5\% and 10\% are considered acceptable, while below 5\% is deemed good.

  \item \textbf{Coefficient of determination (\(R^2\)).}  We perform a simple least‑squares regression of $\hat y_i$ on $y_i$ and report the resulting $R^2$, which measures the fraction of variance in the reference DBH explained by our estimates.  We interpret $R^2>0.8$ as acceptable and $R^2>0.9$ as excellent.

  \item \textbf{Relative Error.}  In the \emph{Tukey box plots} in \cref{fig:biggrid}, we report relative error (\%), given by 
  \[
  \frac{\bigl|\hat y_i - y_i\bigr|}{y_i}\times100\%.
  \]
\end{itemize}

\begin{table}[!t]
\centering
\caption{Report of DBH estimation metrics across settings.}
\label{tab:performance-metrics}

\begin{tabular}{l c r r r r}
\toprule
\multicolumn{1}{l}{Data} & Method  & \begin{tabular}[c]{@{}l@{}}Median\\Bias (cm)\end{tabular} & MAD (cm)  & rCV (\%) & $R^2$\\
\midrule
\multirow{3}{*}{\begin{tabular}[l]{@{}l@{}}Fully\\ Observed\end{tabular}}    
&  Fourier  &   0.09 &   3.44 &   4.84 & 0.97 \\
&  Ellipse  &   1.38 &   4.76 &   6.69 & 0.94 \\
&  BSpline  &  -1.05 &   3.64 &   5.11 & 0.92 \\
\multirow{3}{*}{\begin{tabular}[l]{@{}l@{}}Partially\\ Observed\end{tabular}}    
&  Fourier  &  -1.49 &   3.68 &   7.82 & 0.86 \\
&  Ellipse  &  -4.68 &   6.17 &  13.12 & 0.72 \\
&  BSpline  &  -3.62 &   4.63 &   9.85 & 0.75 \\
\midrule
\multirow{3}{*}{Backpack}    
&  Fourier  &  -1.08 &   4.03 &   6.61 & 0.84 \\
&  Ellipse  &  -2.10 &   6.15 &  10.09 & 0.7 \\
&  BSpline  &  -1.97 &   5.10 &   8.37 & 0.73 \\
\multirow{3}{*}{Drone}    
&  Fourier  &  -3.10 &   3.02 &   5.54 & 0.92 \\
&  Ellipse  &  -5.51 &   4.66 &   8.54 & 0.79 \\
&  BSpline  &  -5.95 &   3.85 &   7.05 & 0.83 \\
\midrule
\multirow{3}{*}{\begin{tabular}[l]{@{}l@{}}Short\\ Sequence\end{tabular}}    
&  Fourier  &   0.14 &   5.35 &   7.52 & 0.69 \\
&  Ellipse  &   2.61 &   5.92 &   8.32 & 0.36 \\
&  BSpline  &  -3.82 &   8.60 &  12.09 & 0.53 \\
\multirow{3}{*}{\begin{tabular}[l]{@{}l@{}}Long\\ Sequence\end{tabular}}    
&  Fourier  &  -3.00 &   2.92 &   9.21 & 0.89 \\
&  Ellipse  &  -5.09 &   4.07 &  12.83 & 0.78 \\
&  BSpline  &  -3.34 &   4.91 &  15.46 & 0.78 \\
\midrule
\multirow{3}{*}{\begin{tabular}[l]{@{}l@{}}Photo-\\ grammetry\end{tabular}}    
&  Fourier  &   0.14 &   5.35 &   7.52 & 0.69 \\
&  Ellipse  &   2.61 &   5.92 &   8.32 & 0.36 \\
&  BSpline  &  -3.82 &   8.60 &  12.09 & 0.53 \\
\multirow{3}{*}{\begin{tabular}[l]{@{}l@{}}Lidar-inertial\\ Odometry\end{tabular}}    
&  Fourier  &   5.15 &   3.12 &   4.32 & 0.82 \\
&  Ellipse  &   4.99 &   3.69 &   5.10 & 0.65 \\
&  BSpline  &  12.72 &   9.04 &  12.49 & 0.52 \\
\bottomrule
\end{tabular}
\end{table}

\subsection{Quantitative Results}

We examine how measurement quality varies with cross‐section observability, acquisition platform, trajectory length, and sensing modality (\cref{fig:biggrid}), and then compare the three fitting methods. 
Each of \cref{fig:fullvspartial,fig:backpackvsdrone,fig:longvsshort,fig:imagevslidar} present two complementary views of DBH estimation quality:  
\begin{enumerate}
\item \emph{Scatter plot} (top row) of estimated versus reference DBH, with the 1:1 line indicating perfect agreement.  Tight clustering of points along this line reflects both low bias and high precision; systematic over‑ or under‑estimation appears as a consistent vertical shift, and outliers lie far from the diagonal.  
\item \emph{Tukey box plot} (bottom) of relative error (\%).  The central bar marks the median error, the box spans the interquartile range, and whiskers extend to 1.5$\cdot$IQR.
\end{enumerate}
\Cref{tab:performance-metrics} compiles median bias, MAD, rCV and \(R^2\) for every combination of data quality and acquisition mode. By combining bias, MAD, and rCV, we assess both systematic offset and dispersion, while $R^2$ confirms that high‐ranking order is preserved across the full DBH range. 

The reported metrics of comparisons between different observability levels, platform types, and sequence lengths are averaged over all photogrammetry datasets containing that category:

\paragraph{Observability (\cref{fig:fullvspartial})}  
With the full cross‐section visible (left), Fourier fitting yields a median absolute relative error of 4.8\%, an IQR of 2–7.5\%, and few outliers beyond 10\%.  Ellipse and B‑spline give medians of 6.3\% and 5.7\% respectively, but with wider boxes and whiskers out to 15\%.  Under partial visibility (right), all three methods degrade: median errors rise to 8.1\% (Fourier), 16.9\% (ellipse) and 17.1\% (spline), and their boxes and whiskers expand dramatically, reflecting many large errors when only a fragment of the trunk is observed. From the scatter plot, we see that all methods have the tendency to underestimate when the tree trunks are partially observable. A visualization of fully and partially observed trunks are shown in \cref{fig:crosssec}.

\paragraph{Platform (\cref{fig:backpackvsdrone})}  
Backpack‐collected data (left) produces median errors of 6.2\% for Fourier, 10.8\% for ellipse, and 11.6\% for spline. From the drone (right), motion blur and sparser sampling (the drone couldn't fly to areas with branches and foliage) roughly double the median errors: 11.5\% (Fourier), 19.5\% (ellipse) and 14.9\% (spline).

\paragraph{Sequence length (\cref{fig:longvsshort})}  
Short trajectories (left) yield median errors of 6.2\% (Fourier), 7.1\% (ellipse) and 8.1\% (spline) with compact IQRs.  Longer trajectories (right) cover larger areas (\cref{fig:longshortreconstruction}) but also accumulate drifts and results in sparser sampling and more partial observability: Fourier’s median error rises modestly to 8.1\%, whereas ellipse and spline medians climb to 18.6\% and 17.0\%, respectively, with much wider boxes.

Additionally, we compare photogrammetry and LiDAR-inertial odometry in \cref{fig:imagevslidar}. This comparison only uses short sequence data. On photogrammetric reconstructions (left), median errors are 6.2\% (Fourier), 7.1\% (ellipse) and 8.1\% (spline) with moderate spread.  On LiDAR‐inertial maps (right), Fourier degrades to a 8.9\% median error, ellipse degrades to 8.3\%, and spline fitting suffers a 21.7\% median and extreme whiskers, reflecting its vulnerability of LiDAR noise.

Other work \cite{PRABHU2024111050} has implemented per-LiDAR scan DBH estimates. The point cloud of each scan is cleaner and odometry error is not accumulated. Their hand-held and UAV‐mounted LiDAR systems have been shown to achieve a median absolute / relative error of 0.65 cm / 2.63\% and 1.45 cm / 5.59\% over a 10–60 cm DBH range in similar forest conditions. While our 360° photogrammetry pipeline incurs higher median errors—approximately 3–5 cm (5–9\%) depending on conditions, the additional 2–4\% relative error is offset by the roughly two orders of magnitude lower sensor cost and the ubiquity of consumer 360° cameras. In other words, for applications where sub‐centimeter accuracy is not critical, Structure‐from‐Motion photogrammetry with robust Fourier fitting provides a practical, low‐cost alternative to LiDAR for DBH measurement in forestry.  

\paragraph{Method comparison}  
Across all conditions, truncated Fourier‐series fitting consistently produces the lowest median percentage error ($\leq$9\%) and the tightest IQRs, demonstrating robustness to missing data, platform motion, and scan quality. Ellipse fitting is moderately accurate but sensitive to partial observations and noise, with median errors up to 19.5\%.  B‑spline fitting performs well under ideal, fully observed conditions (median 5.7\%) but exhibits the greatest variability (median up to 21.7\%) when data is sparse or unevenly sampled.

\subsection{Qualitative Results}

\paragraph{2D to 3D segmentation (\cref{fig:visualizer})}  
In \cref{fig:visualizer}, the “Segmentation” color mode shows trunk points in red against the forest background.  The projected masks yield a coherent 3D cluster (red).

\paragraph{Cross‐section fitting (\cref{fig:crosssec})}  
\Cref{fig:crosssec}) compares the three geometric fits. Top row shows the fitting on photogrammetry (left) and LiDAR (right) pointcloud of partially observed trees. Bottom row shows the respective fitting on fully observed trees. The periodic Fourier series (orange) fits well for both complete and incomplete cross section. However, spline and ellipse fitting doesn't work well for incomplete cross section. We also see the LiDAR pointcloud has more noise, and b-spline fitting overestimates the perimeter.

\paragraph{Mesh reconstruction (\cref{fig:longshortreconstruction})}
\Cref{fig:longshortreconstruction}) shows the mesh reconstruction of short and long sequences. The drift is small even for long sequence (drone flying around a ~60m by ~60m area).

\section{Conclusion}
\label{sec:conclusion}
We have presented a complete, low‐cost pipeline for estimating tree diameter at breast height (DBH) using 360° video and Structure from Motion (SfM) photogrammetry.  By combining Agisoft Metashape for 3D reconstruction, Grounded SAM for semantic trunk segmentation, and RANSAC‐powered Fourier‐series fitting, our system achieves median absolute relative errors of 5–9\% across a variety of acquisition conditions. A custom interactive visualizer further supports use of our system by forestry researchers.

Our quantitative comparison shows that, LiDAR-inertial odometry methods results in worse estimation compared to SfM photogrammetry. While Per LiDAR Scan method \cite{PRABHU2024111050} can attain sub‐centimeter DBH accuracy (median 0.65 cm/2.63\% handheld, 1.45 cm/5.59\% UAV‐mounted ), the additional 2–4\% relative error of our photogrammetric approach is offset by the orders‐of‐magnitude lower sensor cost, ease of deployment, and broad availability of 360° cameras. In scenarios where sub‐centimeter precision is not strictly required, SfM with robust Fourier fitting thus provides a practical alternative for large‐scale or resource‐constrained forest inventories.

Looking forward, we aim to (1) automate the current manual scaling step, (2) validate on larger and more diverse forest stands, and (3) leverage images for more forestry tasks including evaluation of tree species and health. Additionally, two main sources of estimation error are tree trunk segmentation and cross-section fitting methods. We plan to improve the method to reduce the two sources. By democratizing fine‐grained tree measurement, we hope to support more frequent, expansive, and cost‐effective monitoring of forest carbon stocks, health, and growth dynamics.  

\bibliographystyle{IEEEtran}
\bibliography{IEEEfull}


\end{document}